\documentclass{article}

\PassOptionsToPackage{numbers, sort&compress}{natbib}
\PassOptionsToPackage{pagebackref=true,colorlinks,allcolors=blue}{hyperref}
 \usepackage[preprint]{neurips_2026}

\usepackage[utf8]{inputenc} 
\usepackage[T1]{fontenc}    
\usepackage[pagebackref=true,breaklinks,colorlinks,citecolor=blue,linkcolor=blue,urlcolor=blue,hypertexnames=false]{hyperref}
\usepackage{url}            
\usepackage{booktabs}       
\usepackage{amsfonts}       
\usepackage{nicefrac}       
\usepackage{microtype}      
\usepackage{xcolor}         

\usepackage{multirow}
\usepackage{multicol}
\usepackage{adjustbox}
\usepackage[normalem]{ulem}
\useunder{\uline}{\ul}{}
\usepackage{graphicx}
\usepackage{subcaption}
\usepackage{booktabs}
\usepackage{amsmath}
\usepackage{tabularx}
\usepackage{wrapfig}
\usepackage{xspace}
\usepackage{listings}
\usepackage[font=small]{caption}
\usepackage[linesnumbered,ruled,vlined]{algorithm2e}
\SetKwInput{KwRequire}{Require}
\SetKwInput{KwHyperp}{Hyperparameters}
\usepackage{caption}
\usepackage{float}
\usepackage{wrapfig}
\usepackage{enumitem}
\usepackage[capitalize]{cleveref}
    \crefname{section}{Sec.}{Secs.}
    \Crefname{section}{Section}{Sections}
    \Crefname{table}{Table}{Tables}
    \crefname{table}{Table}{Tables} 

\usepackage{amsmath,amsfonts,bm}

\def\1{\bm{1}}

\DeclareMathAlphabet{\mathsfit}{\encodingdefault}{\sfdefault}{m}{sl}
\SetMathAlphabet{\mathsfit}{bold}{\encodingdefault}{\sfdefault}{bx}{n}

\newcommand{\ours}{\textsc{VAORA}\xspace}

\title{Bridging Physical Reasoning and Task Generalization 
via Visual Action Outcome Reasoning Alignment}

\author{%
  Han-Jun Ko$^{*1}$ \quad Jr-Jen Chen$^{*1}$ \quad Haobo Yuan$^{2}$ \quad
  Hsin-Ying Lee$^{2}$ \quad Tiancheng Shen$^{2}$ \\[0.3em]
  \bfseries
  Ming-Hsuan Yang$^{2}$ \quad Yu-Chiang Frank Wang$^{1}$ \\[0.5em]
  \normalfont\normalsize
  $^{1}$National Taiwan University \qquad
  $^{2}$The University of California, Merced
}

\begin{document}

\maketitle

\begin{abstract}
Vision-language models (VLMs) struggle to generalize in interactive physical reasoning, particularly under unseen tasks and environments. Two key failure modes are prominent: hallucinated chain-of-thought (CoT) reasoning that contradicts physical reality, and misalignment between the model’s reasoning and actions. We present VAORA (\textit{Visual Action Outcome Reasoning Alignment}), a novel reward design that directly addresses both issues. VAORA introduces two complementary rewards: \textit{Visual Alignment Reward}, which anchors VLM reasoning to the visual context independent of the agent action itself, and \textit{Visual-Action Alignment Reward}, which grounds reasoning in the visual outcome induced by the model’s action. Together, these rewards suppress hallucinated CoT and reduce the gap between reasoning and behavior. To improve training stability, we further employ smooth, dense rewards by estimating success probabilities using a pre-trained in-domain expert agent. Experiments on PHYRE and Virtual Tool support our performances across novel-task and unseen-environment settings, confirming that grounded and generalizable physical intelligence can be induced through VAORA.
\end{abstract}

\begingroup
\renewcommand\thefootnote{}
\footnotetext{$^{*}$ denotes co-first author.}
\addtocounter{footnote}{-1}
\endgroup

\section{Introduction}
\label{sec:intro}

True physical agents must go beyond memorizing scene configurations and instead reason about spatial relationships, dynamics, and causality to act effectively in novel situations. We characterize this capability through two forms of generalization: cross-task transfer within the same environment and cross-environment transfer across distinct physics simulators.
Conventional non-vision-language-model agents are fundamentally constrained in satisfying these criteria by their architectural design, which typically combines a visual encoder such as Vision Transformer (ViT)~\citep{dosovitskiy2021an} or ResNet18~\citep{he2016deep} with a Multi-Layer Perceptron (MLP) for direct action prediction~\citep{bakhtin2019phyre, girdhar2020forward, qi2021learning, ahmed2021dynamics, li2022learning}.
With perception and decision-making collapsed into a single opaque mapping, these architectures lack interpretable intermediate representations, making it difficult to explain or audit their decisions. Moreover, prior work~\citep{langosco2022goal, delfosse2025deep, cobbe2019quantifying, geirhos2020shortcut} shows that direct visual-to-action learning often exploits spurious correlations rather than transferable physical principles. As a result, agents trained on benchmarks such as PHYRE~\citep{bakhtin2019phyre} exhibit brittle behavior under distributional shifts: strong performance on seen tasks comes at the expense of generalization.

\begin{figure}[!t]
  \centering
  \includegraphics[width=0.92\textwidth]{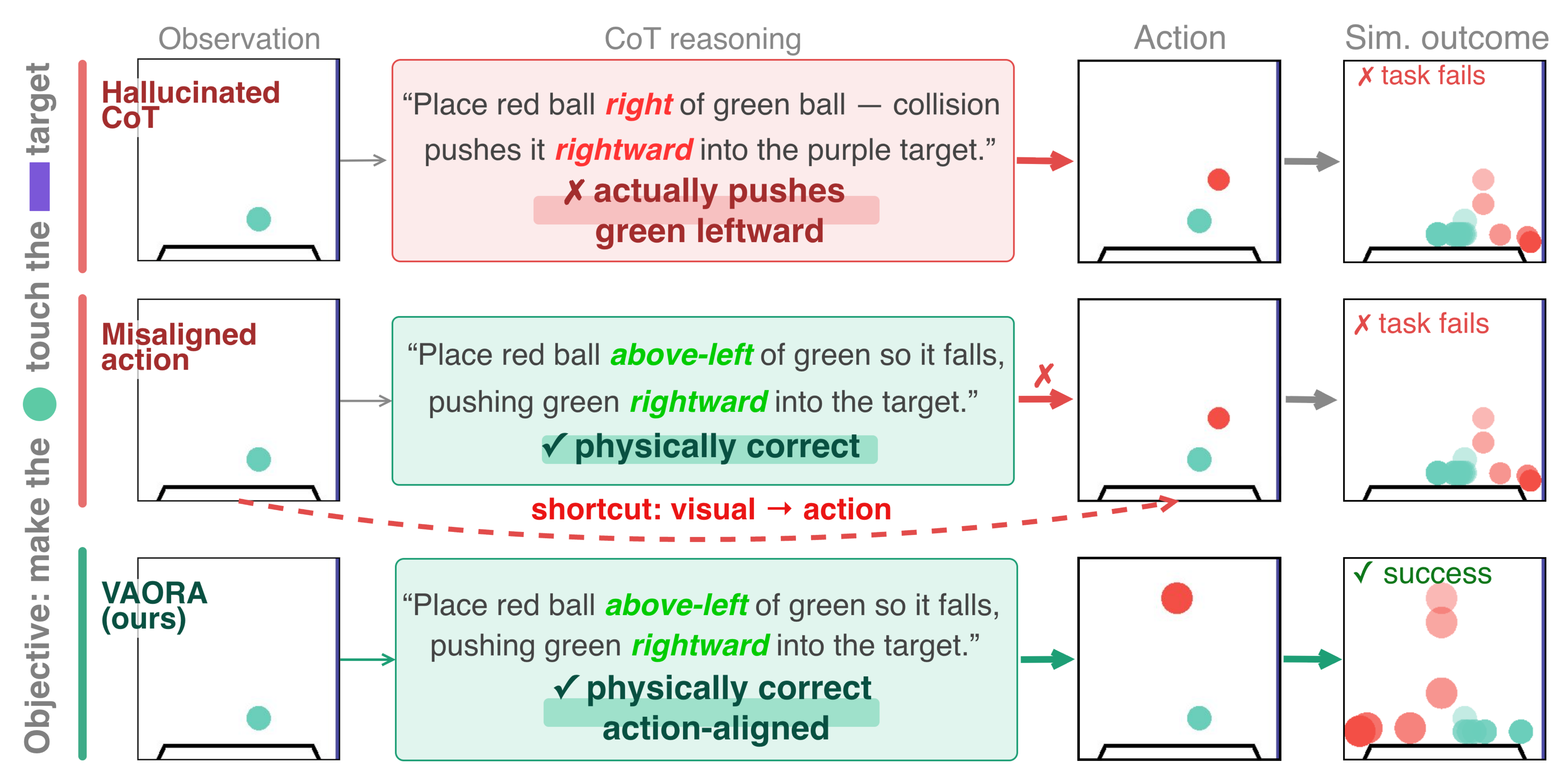}
  \caption{\textbf{Two major obstacles in CoT-based physical reasoning.} Hallucinated CoT denotes the model producing physically incorrect reasoning that leads to wrong action; Misaligned Action, on the other hand, bypasses physically-aligned reasoning via a visual shortcut and results in wrong action planning. Our VAORA aims to resolve both issues, generating successful action with proper physical reasoning.}
  \label{fig:teaser}
  \vspace{-4mm}
\end{figure}

Vision-language models (VLMs)~\citep{qwen2025qwen25, dubey2024llama3} offer a fundamentally different paradigm by replacing reactive mappings with explicit causal reasoning. Through chain-of-thought (CoT) reasoning~\citep{wei2022chain, kojima2022zero, wang2022self}, VLMs can form causal interpretations of physical dynamics, enabling stronger cross-task and cross-environment generalization~\citep{zawalski2024ecot, zhao2025cotvla}. However, the two dominant training paradigms, Supervised Fine-Tuning (SFT)~\citep{chen2025sft} and reinforcement learning optimized solely for task success, each introduce structural limitations that hinder this potential~\citep{li2025diagnosis, li2025temporalhalluc}.

Prior work shows that SFT primarily teaches models to imitate the linguistic form of expert reasoning without grounding the reasoning process in physical reality~\citep{chen2025sft, ross2011reduction, lanham2023measuring, turpin2023language}. In contrast, success-driven RL often encourages models to bypass CoT reasoning entirely, reverting to shortcut-driven visual-to-action mappings that exploit dataset biases rather than learning transferable physical principles~\citep{shao2025spurious, wen2025ends, visionary2025, cobbe2019quantifying, geirhos2020shortcut}. Consequently, both approaches fail to establish explicit supervision between CoT reasoning and physical reality, leaving critical failure cases unresolved, as depicted in \cref{fig:teaser}.

To address both failure modes, we introduce \textbf{VAORA} (\textit{Visual Action Outcome Reasoning Alignment}), a novel reward design for cultivating grounded and generalizable physical reasoning in VLMs. VAORA consists of two complementary reward signals. The \textit{Visual Alignment Reward} anchors the reasoning process to action-independent visual context, suppressing hallucinated reasoning at its source. The \textit{Visual-Action Alignment Reward} grounds the reasoning trace in the visual outcome induced by the model’s action, jointly mitigating hallucinated CoT and reducing the gap between reasoning and behavior.
To further stabilize training in continuous-action interactive physical reasoning, where rewards are naturally sparse and noisy, we augment these signals with smooth and dense success-probability estimates derived from a pre-trained in-domain expert agent.

We conduct extensive experiments across multiple physical reasoning benchmarks to evaluate the generalization capability of our approach. Specifically, we assess cross-\textit{task} generalization by training on a subset of PHYRE~\citep{bakhtin2019phyre} configurations and evaluating on held-out tasks.
We further evaluate cross-\textit{environment} generalization through zero-shot transfer from PHYRE to Virtual Tool~\citep{allen2020rapid}. In addition, experiments on the Craft Visual Question Answering (VQA) benchmark~\citep{ates2020craft} demonstrate that the improved physical reasoning induced by our method transfers beyond action selection to explicit reasoning and question answering.
The main contributions of this work are:
\begin{itemize}[leftmargin=*, topsep=0pt, nosep]
    \item \textbf{Diagnostic Insight:} We identify that both SFT and success-driven RL fail to supervise the connection between CoT reasoning and physical reality, leading to two fundamental failure modes: hallucinated chain-of-thought (CoT) reasoning that contradicts physical reality, and misalignment between the model’s stated reasoning and its executed actions.
    \item \textbf{Dual Alignment Reward Design:} To address the above challenges, we introduce VAORA, a novel reward framework consisting of two complementary reward signals: \textit{Visual Alignment Reward}, which anchors reasoning to action-independent visual context, and \textit{Visual-Action Alignment Reward}, which grounds reasoning in the visual consequences of the model’s actions.
    \item \textbf{From Physical Reasoning to Task Generalization:} We demonstrate cross-task generalization by surpassing the DQN expert on unseen PHYRE tasks~\citep{bakhtin2019phyre}, while achieving zero-shot cross-environment transfer to Virtual Tool~\citep{allen2020rapid}, matching or outperforming frontier closed-source models. These generalization gains are further supported by deeper physical understanding beyond action selection on the Craft VQA benchmark~\citep{ates2020craft}.
\end{itemize}
\raggedbottom
\vspace{-5pt}
\section{Related Work}
\vspace{-5pt}

\textbf{Interactive Physical Reasoning.}
The paradigm for interactive physical reasoning relies on task-success-driven RL agents, with non-VLM-based, including policy and value-based methods~\citep{bakhtin2019phyre, girdhar2020forward, qi2021learning, ahmed2021dynamics, li2022learning}. Although these approaches achieve strong performance on seen tasks, their architectural design inherently limits their capacity for broader generalization. In contrast, VLM-based methods have shown substantial progress in passive physical reasoning~\citep{ates2020craft, yi2020clevrer, chow2025physbench} and exhibit strong generalizability. However, results on DeepPHY~\citep{xu2025deepphy} indicate that their performance in interactive environments remains far from satisfactory.
In this work, we bridge this gap by introducing visually grounded rewards into VLM training to enhance generalizability and by stabilizing training through learning RL agent predictions.

\textbf{Training Paradigms for VLM Physical Reasoning.}
Existing approaches to physical reasoning in VLMs broadly fall into two categories: Supervised Fine-Tuning (SFT)~\citep{chen2025sft, ross2011reduction} and RL with success-driven rewards~\citep{shao2024deepseekmath, guo2025deepseekr1}. SFT distills reasoning capabilities from frontier closed-source models, encouraging VLMs to imitate expert reasoning traces. However, without grounding reasoning in observable physical outcomes, SFT often produces fluent yet physically inconsistent chains of thought~\citep{lanham2023measuring, turpin2023language}. 
In contrast, success-driven RL directly optimizes task completion, but often encourages shortcut behaviors that bypass CoT reasoning and exploit dataset-specific correlations instead of learning causal understanding~\citep{shao2025spurious, wen2025ends, visionary2025}. Both paradigms fail to supervise the alignment between reasoning traces and physical reality, limiting the generalization of physical reasoning~\citep{bagdonaviciute2025physics}.

\textbf{Outcome Prediction as Physical Supervision.}
Recognizing that models without grounding in physical reality struggle to reason causally, prior work has explored incorporating environmental feedback through several directions: learning object-dynamics models~\citep{battaglia2016interaction, fragkiadaki2016billiards}, predicting future states to rank candidate plans~\citep{girdhar2020forward, qi2021learning}, and leveraging external tools such as language-grounded simulators~\citep{zellers2021piglet} or text-to-video generators~\citep{du2023vlp}. Despite their diversity, these approaches all depend on separate dynamics modules or generation tools as intermediaries, rather than directly incorporating post-interaction outcomes into VLM reasoning end-to-end.

\textbf{Reinforcement Learning for VLM Reasoning.}
Applying RL to train VLMs has gained significant attention through approaches such as RLHF~\citep{ouyang2022training} and GRPO~\citep{shao2024deepseekmath}, which provide policy-gradient-based optimization for reasoning alignment. However, prior work consistently shows that optimization under sparse and noisy rewards can lead to severe training instability and model collapse~\citep{shao2025spurious, wen2025ends, visionary2025, zhao2026robustness, castanyer2025stabilizing}. Motivated by these findings, we leverage dense success-probability estimates derived from a pre-trained expert agent to stabilize optimization and complement VAORA's reward design.
\vspace{-5pt}
\section{Method}
\vspace{-5pt}
\label{sec:method}

\begin{figure}[t]
  \centering
  \includegraphics[width=\textwidth]{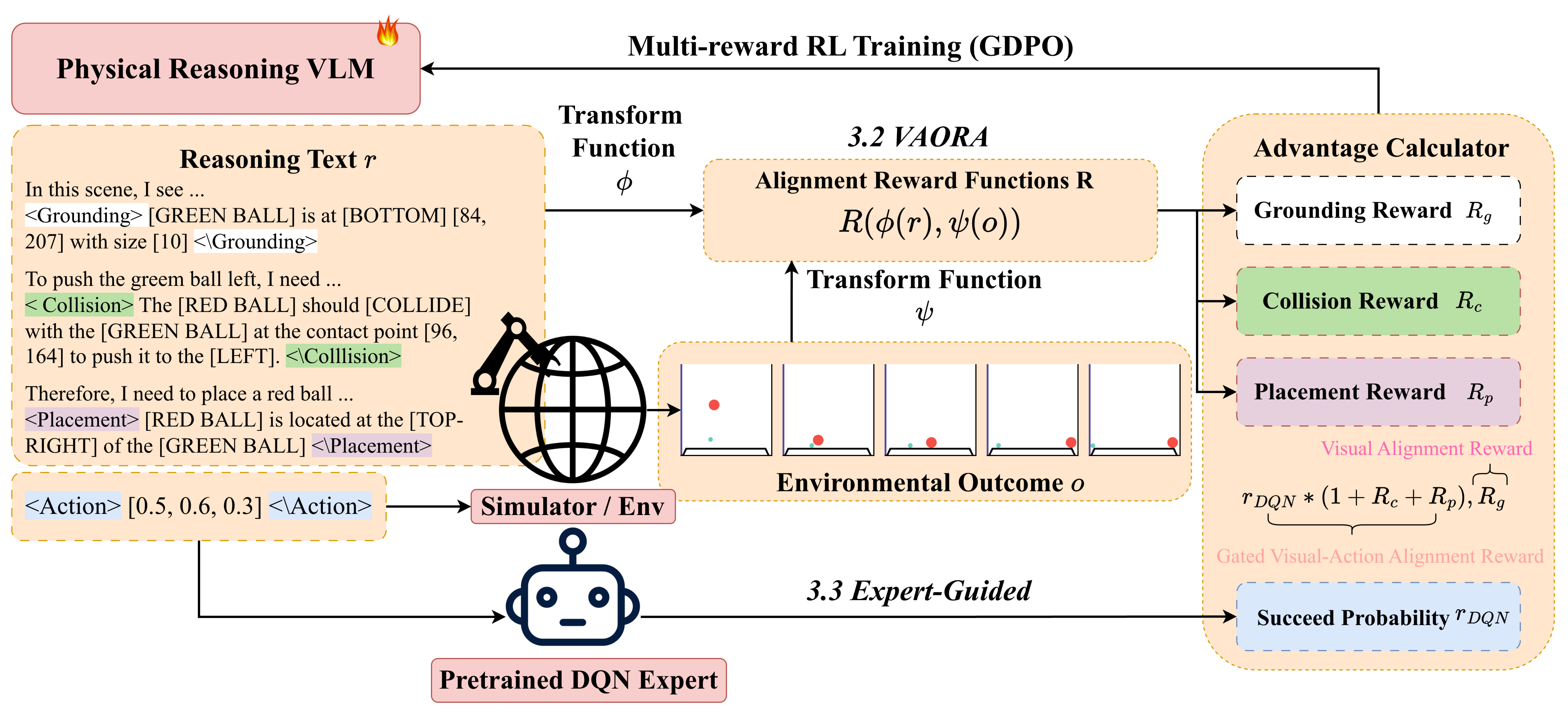}
  \caption{\textbf{Framework.} \ours consists of two components: (a) Visual-Alignment Reward, which anchors reasoning to action-independent visual context. (b) Gated Visual-Action Alignment Reward, which aligns reasoning with the visual outcome of the model's actions, gated by a success probability estimated by an expert model.}
  \label{fig:main}
  \vspace{-4mm}
\end{figure}
We present \textbf{VAORA}, a novel reward design that directly addresses two fundamental failure modes limiting the generalization of VLMs in interactive physical reasoning: \textit{hallucinated CoT reasoning} that contradicts physical reality, and \textit{misalignment} between the model's stated reasoning and its actual actions. 
An overview of the pipeline is shown in~\cref{fig:main}.

\subsection{Problem Formulation}

Interactive physical reasoning tasks require an agent to intervene in a physical scene, \emph{e.g.}, by placing or launching objects such that the resulting dynamics satisfy a specified goal. PHYRE~\citep{bakhtin2019phyre} instantiates this challenge as a 2D physics environment in which the agent must determine a single continuous action that triggers a sequence of physical interactions leading to task success.
 
Formally, given an initial scene observation $\mathbf{o} \in \mathcal{O}$ consisting of a visual depiction of the physical setup and a natural language task description, the model is required to generate a chain-of-thought reasoning trace $\mathbf{r}$ followed by a single continuous action:
\begin{equation}
    \mathbf{a} = [x,\, y,\, r] \;\in\; \mathcal{A} \subset \mathbb{R}^3,
\end{equation}
where $x, y \in [0, 1]$ denote normalized placement coordinates and $r \in [0, 1]$ denotes the normalized radius of the intervening ball. The environment executes $\mathbf{a}$ and returns a post-interaction outcome $\mathbf{v}$ together with a binary task-success signal $\mathcal{S} \in \{0, 1\}$.
\subsection{Visual Action Outcome Reasoning Alignment (VAORA)}
\label{sec:method-vora}

Both failure modes identified in~\cref{sec:intro}, namely \textit{hallucinated CoT reasoning} and reasoning-action \textit{misalignment}, stem from a common issue: the model's reasoning trace is never supervised against the actual outcome of its intervention in the environment. VAORA addresses both through a unified alignment principle requiring the reasoning trace to remain consistent with the \textit{visual observations of the environment}.

Specifically, we align the reasoning chain with both the initial scene observation and the post-interaction environmental outcome within a shared symbolic space. Formally, let $\mathbf{r}$ denote the model's reasoning chain, $\mathbf{o}$ the initial scene observation, and $\mathbf{v}$ the post-interaction environmental outcome. We first apply domain-specific transformations $\phi(\cdot)$ and $\psi(\cdot)$ to project the reasoning trace and visual observations into a shared symbolic space:
\begin{equation}
    r_{\mathrm{VAORA}} = \mathcal{R}\!\left(\phi(\mathbf{r}),\, \psi(\mathbf{o}),\,
    \psi(\mathbf{v})\right),
\end{equation}
where $\phi(\mathbf{r})$ extracts structured symbolic propositions from the reasoning trace, $\psi(\mathbf{o})$ and $\psi(\mathbf{v})$ derive ground-truth symbolic states from the initial scene and post-interaction outcome, respectively, and $\mathcal{R}(\cdot,\cdot,\cdot)$ measures their consistency. The formulation of $\phi$ and $\psi$ is general and can be instantiated for any domain in which a reasoning trace makes verifiable predictions about observable outcomes, such as spatial relationships in robotic manipulation
or fluid flow directions in pipe-routing puzzles.
In our physical reasoning setting, $\phi$ parses structured symbolic propositions from the model's reasoning trace, and $\psi$ extracts ground-truth physical states from the simulator.

With symbolic representations of reasoning, scene observation, and outcomes, we propose three rewards under two categories to enable alignment:

\textbf{Visual Alignment Reward.}
To anchor the reasoning trace to a scene independent of the model's action, we design \textbf{grounding reward} $r_{\mathrm{G}}$, which aligns the VLM's perception of initial scene configuration with the real initial scene observation $\mathbf{o}$, directly addressing hallucinated CoT reasoning.
A model that cannot accurately perceive the scene before acting will inevitably produce reasoning inconsistent with physical reality.
Through prompt engineering (see Appendix~\cref{app:prompt-phyre}), the VLM is encouraged to structure its reasoning trace into a symbolic space $\phi(\mathbf{r})$ by expressing each object $i$ as a symbolic tuple $(\hat{\ell}^{(i)}, \hat{x}^{(i)}, \hat{y}^{(i)}, \hat{s}^{(i)})$ within \texttt{<scene\_answer>} tags, 
where $\hat{\ell}^{(i)}$ denotes a textual spatial label referenced to the full scene, drawn from a predefined $3\times3$ spatial grid of nine regions (e.g., \texttt{TOP-LEFT}, \texttt{CENTER}; see Appendix~\ref{app:vaora} for full definition).

while $(\hat{x}^{(i)}, \hat{y}^{(i)}, \hat{s}^{(i)})$ represent the predicted normalized coordinates and size.
We then extract all objects from the initial scene observation $\mathbf{o}$ and transform them into the same symbolic space $\psi(\mathbf{o})$ to obtain ground-truth tuples $(\ell^{(i)}, x^{(i)}, y^{(i)}, s^{(i)})$, where $\ell^{(i)}$ is assigned by mapping object coordinates to the corresponding canonical region.
 
The grounding reward for the $i$-th object is defined as:
\begin{equation}
    r_{\mathrm{G}}^{(i)} = \frac{1}{2}\,\mathrm{soft}\!\left(\hat{\ell}^{(i)},\,
    \ell^{(i)}\right) + \frac{1}{2}\left(1 - \frac{\|(\hat{x}^{(i)},\hat{y}^{(i)},
    \hat{s}^{(i)}) - (x^{(i)}, y^{(i)}, s^{(i)})\|}{\sqrt{3}}\right),
\end{equation}
where $\mathrm{soft}(\hat{\ell}^{(i)}, \ell^{(i)})$ is a soft grid score measuring the proximity between predicted and ground-truth spatial labels (see Appendix~\cref{app:vaora} for details), bridging low-level coordinate perception and open-vocabulary spatial reasoning. The reward combines two complementary components: a soft spatial grounding term in textual space and a numerical deviation term evaluating predicted coordinates and size. A penalty of $-p_g$ is applied when a predicted object does not exist in the scene, is duplicated, or contains formatting errors. The final grounding reward aggregates per-object scores with a weight of $w_g$ and is clamped to $[0,1]$.

\textbf{Visual-Action Alignment Reward.}
We propose two additional rewards to ground the reasoning trace in the physical consequences of the model's action.
The \textbf{collision reward} $r_C$ aligns the model's predicted collision events with the post-interaction visual outcome, addressing both reasoning-action misalignment and hallucinated CoT. 

With this alignment, the model is rewarded only when its reasoning predicts collision events that correspond to both its action and physical reality.
Through prompt engineering (see Appendix~\cref{app:prompt-phyre}), the VLM is guided to structure its reasoning trace into a symbolic space $\phi(\mathbf{r})$ by expressing each predicted collision event $i$ as a symbolic tuple $(\widehat{\text{obj1}}^{(i)}, \widehat{\text{obj2}}^{(i)}, \widehat{\text{act}}^{(i)}, \hat{c}x^{(i)}, \hat{c}y^{(i)}, \hat{d}^{(i)})$ within \texttt{<causal\_actions\_answer>} tags, where $\widehat{\text{act}}^{(i)} \in \{\texttt{COLLIDE}, \texttt{ROTATE}\}$ is the predicted action type, $(\hat{c}x^{(i)}, \hat{c}y^{(i)})$ is the predicted contact point, and $\hat{d}^{(i)}$ is the predicted motion direction --- $\hat{d}^{(i)} \in \{\texttt{LEFT}, \texttt{RIGHT}, \texttt{UP}, \texttt{DOWN}\}$ for \texttt{COLLIDE} events and $\hat{d}^{(i)} \in \{\texttt{CLOCKWISE}, \texttt{COUNTERCLOCKWISE}\}$ for \texttt{ROTATE} events.
The ground-truth collision events $\psi(\mathbf{v})$ are extracted from the post-interaction simulation outcome, providing the reference symbolic tuples $(\text{obj1}^{(i)}, \text{obj2}^{(i)}, \text{act}^{(i)}, cx^{(i)}, cy^{(i)}, d^{(i)})$ for each detected event.
 
By aligning $\phi(\mathbf{r})$ with $\psi(\mathbf{v})$, the reward for each predicted
collision event $i$ is computed as:
\begin{equation}
    r_{\mathrm{C}}^{(i)} = 1 - \frac{\sqrt{(\hat{c}x^{(i)} - cx^{(i)})^2 +
    (\hat{c}y^{(i)} - cy^{(i)})^2}}{\sqrt{2}},
\end{equation}
when the predicted tuple $(\widehat{\text{obj1}}^{(i)}, \widehat{\text{obj2}}^{(i)},
\widehat{\text{act}}^{(i)}, \hat{d}^{(i)})$ matches the ground-truth event in object
identities, action type, and direction. A penalty of $-p_c$ is applied per unmatched
event, including cases of wrong object identity, action type, direction, or format
errors. The final collision reward accumulates per-event scores, weighted by $w_c$ and
clamped to $[0, 1]$.

The \textbf{placement reward} $r_P$
aligns the model's predicted ball placement with its actual position in the first post-interaction frame $\mathbf{v_0}$,
directly addressing reasoning-action misalignment. The model's reasoning only makes
physical sense when the ball is placed where the model intends.
Through prompt engineering (see Appendix~\cref{app:prompt-phyre}), the VLM is guided to structure its reasoning trace into a symbolic space $\phi(\mathbf{r})$ by expressing each predicted placement description $i$ as a symbolic tuple $(\hat{\ell}^{(i)}, \widehat{\text{obj}}^{(i)})$ within \texttt{<placement\_answer>} tags for the exact format), 
where $\hat{\ell}^{(i)}$ is a textual spatial label drawn from the same $3\times3$ spatial grid of nine regions (see Appendix~\ref{app:vaora}), with the difference that the region is now computed relative to the reference object $\widehat{\text{obj}}^{(i)}$ rather than the whole scene.
The ground-truth placement $\psi(\mathbf{v})$ is extracted from the first frame of the post-interaction outcome, capturing the spatial relationship between the placed ball and all other objects, and transformed into reference symbolic tuples $(\ell^{(i)}, \text{obj}^{(i)})$.
 
By aligning $\phi(\mathbf{r})$ with $\psi(\mathbf{v})$, the reward for each predicted
placement description $i$ is computed as:
\begin{equation}
    r_{\mathrm{P}}^{(i)} = \mathrm{soft}\!\left(\hat{\ell}^{(i)},\, \ell^{(i)}\right),
\end{equation}
when the predicted object $\widehat{\text{obj}}^{(i)}$ matches the ground-truth object
$\text{obj}^{(i)}$,where $\mathrm{soft}(\hat{\ell}^{(i)}, \ell^{(i)})$ is a soft grid score that measures
the proximity between the predicted and ground-truth spatial labels (see
Appendix~\cref{app:vaora} for details), which bridges open-vocabulary reasoning back to the
low-level coordinate space required for precise action control. A penalty of $-p_p$ is
applied per description when the predicted object does not match, or a format error is
detected. The final placement reward accumulates per-description scores, weighted by $w_p$ and clamped to $[0, 1]$.

\subsection{Expert-Guided (EG) Policy Optimization and Reward Composition}

Sparse binary task-success rewards in continuous action spaces often cause policy collapse during VLM training~\citep{shao2025spurious, wen2025ends, visionary2025}. To mitigate this issue, we replace the binary environment success signal with a smooth and dense success probability estimated by a pre-trained DQN expert:
\begin{equation}
    r_{\mathrm{DQN}}(\mathbf{a}) = \hat{p}_{\mathrm{DQN}}(\mathbf{o},\, \mathbf{a}),
\end{equation}
where $\hat{p}_{\mathrm{DQN}}(\mathbf{o}, \mathbf{a})$ estimates the probability of task success for action $\mathbf{a}$ under the initial scene $\mathbf{o}$. Additional training details of the DQN expert are provided in Appendix~\cref{app:dqn}.

The final reward integrates all VAORA components together with $r_{\mathrm{DQN}}$.
Since $r_{\mathrm{P}}$ and $r_{\mathrm{C}}$ are visual-action alignment rewards that evaluate the physical consequences of the model's action, they are gated by $r_{\mathrm{DQN}}$ to ensure that reasoning quality is rewarded only when the action itself is physically plausible.
In contrast, $r_{\mathrm{G}}$, as a visual-alignment reward independent of the action, is optimized outside the gate.
We then jointly optimize the two reward components using GDPO~\citep{liu2026gdpo}: the \textit{gated visual-action alignment reward}: $r_{\mathrm{DQN}} \cdot (1 + r_{\mathrm{P}} + r_{\mathrm{C}})$ and the \textit{visual-alignment reward}: $r_{\mathrm{G}}$.

\section{Experiments}
\label{sec:exp}

\begin{figure}[t]
  \centering
  \includegraphics[width=\textwidth]{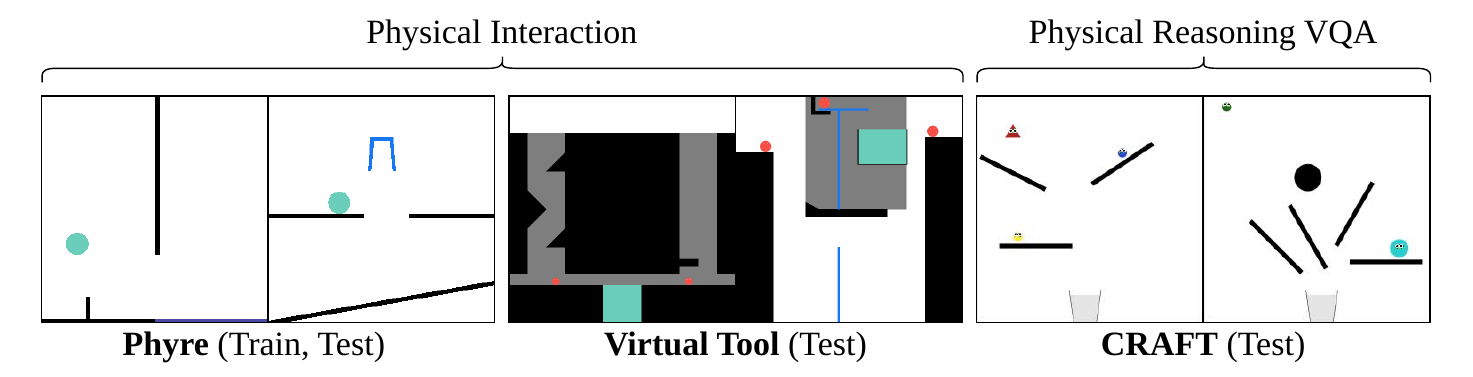}
  \caption{\textbf{Generalization of VAORA.} We train \ours on Phyre tasks and evaluate across three scenarios: (a) \textbf{Phyre} test set for cross-task generalization, (b) \textbf{Virtual-Tool} for cross-engine generalization, and (c) \textbf{CRAFT VQA} to assess the transferability of learned reasoning to visual question answering.}
  \label{fig:benchmarks}
  \vspace{-4mm}
\end{figure}

We evaluate VAORA across three benchmarks that probe physical reasoning at multiple levels of generalization, as demonstrated in~\cref{fig:benchmarks}: \textbf{PHYRE}~\citep{bakhtin2019phyre} for cross-task generalization to unseen physical tasks, \textbf{Virtual Tool}~\citep{allen2020rapid} for cross-environment transfer to an entirely different physics simulator, and \textbf{CRAFT VQA}~\citep{ates2020craft} for causal and counterfactual physical understanding. Detailed descriptions and evaluation protocol adaptations are provided in Appendix~\ref{app:benchmarks}.

\subsection{Experimental Setup}

\textbf{Baselines.}
We compare against open-source VLMs including InternVL-3.5-8B~\citep{zhu2025internvl3} and Qwen3-VL-8B-Instruct~\citep{bai2025qwen3vl} under zero-shot, SFT, GRPO, and SFT+GRPO settings; closed-source API models including Claude Sonnet-4.6~\citep{anthropic2026claudesonnet46}, GPT-5.4~\citep{openai2026gpt54}, Gemini-3.1-Flash~\citep{google2026gemini31flash}, and Gemini-3.1-Pro~\citep{google2026gemini31pro}; and the pretrained \textbf{DQN expert}~\citep{mnih2015humanlevel} used as expert guidance.

\textbf{Metrics.}
On PHYRE, we report Pass@$k$ ($k \in \{1, 3, 5\}$), measuring the fraction of tasks 
solved within $k$ attempts. On Virtual Tool, since models are permitted to attempt every available tool on each task, a task is considered solved at the minimum number of attempts required across all tools; we report Pass@$k$ under this definition. On CRAFT VQA, we report accuracy across five question categories and their overall average.

\textbf{Implementation Details.}
We use Qwen3-VL-8B-Instruct as the base model and supervise fine-tuning it on a Gemini-3.1-Flash-generated reasoning trace dataset before RL training. For cross-task generalization on PHYRE, we split the 25 task types into disjoint training and testing sets, conducting three different split configurations to demonstrate the stability of VAORA across different held-out task distributions. For cross-environment generalization on Virtual Tool and broader physical understanding on CRAFT VQA, the model is trained on all 25 PHYRE task types without any held-out split. All ablation experiments share the same base model checkpoint for fair comparison. Full details of the implementation are provided in Appendix~\ref{app:implementation}.

\subsection{Cross-tasks Generalization on PHYRE}

\begin{table*}[t]
\small
\centering
\newcolumntype{C}{>{\centering\arraybackslash}X}
\newcolumntype{L}{>{\raggedright\arraybackslash}X}
\newcolumntype{R}{>{\raggedleft\arraybackslash}X}
\begin{tabularx}{\textwidth}{LLCCCCCCCCC}
\toprule
\multicolumn{2}{c}{} & \multicolumn{3}{c}{Testing Set 1} & \multicolumn{3}{c}{Testing Set 2} & \multicolumn{3}{c}{Testing Set 3} \\
\cmidrule(lr){3-5} \cmidrule(lr){6-8} \cmidrule(lr){9-11}
\multicolumn{2}{c}{} & Pass@1 & Pass@3 & Pass@5 & Pass@1 & Pass@3 & Pass@5 & Pass@1 & Pass@3 & Pass@5 \\
\midrule
\multicolumn{11}{l}{\textit{Oracle}} \\
\midrule
\multicolumn{2}{l}{Human} & 0.520 & 0.758 & 0.800 & 0.472 & 0.822 & 0.920 & 0.520 & 0.732 & 0.820 \\
\multicolumn{2}{l}{DQN-expert} & 0.326 & 0.434 & 0.476 & \textbf{0.260} & 0.368 & 0.418 & \textbf{0.338} & 0.412 & 0.466 \\
\midrule
\multicolumn{11}{l}{\textit{Open-Sourced Baselines}} \\
\midrule
\multicolumn{2}{l}{InternVL-3.5-8B} & 0.044 & 0.094 & 0.134 & 0.028 & 0.084 & 0.130 & 0.050 & 0.104 & 0.154 \\
\multicolumn{2}{l}{Qwen3VL-8B} & 0.002 & 0.008 & 0.200 & 0.018 & 0.044 & 0.078 & 0.016 & 0.040 & 0.074 \\
\multicolumn{2}{l}{Qwen3VL-8B (SFT)} & 0.176 & 0.380 & 0.470 & 0.146 & 0.344 & \underline{0.452} & 0.176 & 0.368 & 0.452 \\
\multicolumn{2}{l}{Qwen3VL-8B (GRPO)} & 0.006 & 0.040 & 0.062 & 0.016 & 0.038 & 0.056 & 0.024 & 0.06 & 0.108 \\
\multicolumn{2}{l}{Qwen3VL-8B (SFT+GRPO)} & 0.020 & 0.026 & 0.030 & 0.042 & 0.124 & 0.172 & 0.022 & 0.078 & 0.102 \\
\midrule
\multicolumn{11}{l}{\textit{Closed-Source API}} \\
\midrule
\multicolumn{2}{l}{Claude-Sonnet-4.6} & 0.042 & 0.104 & 0.142 & 0.036 & 0.088 & 0.122 & 0.030 & 0.074 & 0.108 \\
\multicolumn{2}{l}{GPT-5.4} & 0.064 & 0.132 & 0.186 & 0.048 & 0.096 & 0.144 & 0.044 & 0.086 & 0.140 \\
\multicolumn{2}{l}{Gemini-3.1-Pro} & 0.278 & \underline{0.472} & \underline{0.552} & \underline{0.220} & \underline{0.368} & 0.420 & 0.278 & \textbf{0.484} & \textbf{0.554} \\
\multicolumn{2}{l}{Gemini-3.1-Flash} & 0.170 & 0.340 & 0.416 & 0.192 & 0.302 & 0.352 & 0.142 & 0.294 & 0.376 \\
\midrule
\multicolumn{11}{l}{\textit{\textbf{Ours (base model: Qwen3VL-8B-SFT)}}} \\
\midrule
\multicolumn{2}{l}{+EG (Expert-Guided)} & 0.270 & 0.436 & 0.512 & 0.178 & 0.312 & 0.362 & 0.218 & 0.340 & 0.392 \\
\multicolumn{2}{l}{+EG+VAORA} & \textbf{0.382} & \textbf{0.524} & \textbf{0.594} & 0.198 & \textbf{0.39} & \textbf{0.472} & \underline{0.300} & \underline{0.456} & \underline{0.512} \\
\bottomrule
\end{tabularx}
\caption{\textbf{Results on Phyre test sets.} \ours outperforms all VLM baselines across all three testing sets and surpasses the in-domain DQN expert on the majority of evaluation metrics, demonstrating strong cross-task generalization. \textbf{Bold} and \underline{underline} denote the best and second-best results(excluding human), respectively.}
\label{tab:phyre_main}
\vspace{-6mm}
\end{table*}

As shown in~\cref{tab:phyre_main}, \ours substantially outperforms all open-source VLMs and the majority of closed-source API baselines across every testing set and all $k$ values. Against Gemini-3.1-Pro, a much larger frontier model, \ours surpasses it on Testing Sets~1 and~2 and remains competitive on Testing Set~3. We note that, \ours surpasses the DQN agent, the very source of its expert reward signal, across nearly all metrics on all three testing sets. This demonstrates that, with properly aligned reasoning via VAORA, the VLM's capacity for symbolic physical reasoning confers a genuine generalization advantage over direct visual-to-action mapping agents such as DQN on unseen tasks.

It is worth noting that, applying RL with only the native binary task-success signal, as in the (GRPO) and (SFT+GRPO) baselines, leads to severely degradation relative to the SFT baseline, providing direct empirical evidence of training instability under sparse rewards in continuous action spaces~\citep{andrychowicz2017hindsight, shao2025spurious, wen2025ends}. The Expert-Guided (EG) reward replaces this binary signal with the DQN expert's smooth success-probability estimates, yielding stable training dynamics and consistent improvement over the SFT baseline across all testing sets. And, comparing +EG against +EG+VAORA in~\cref{tab:phyre_main}, our full version consistently improves Pass@1, Pass@3, and Pass@5 across all three testing sets.

\subsection{Cross-Environment Generalization on Virtual Tool}
\begin{table*}[t]
\small
\centering
\newcolumntype{C}{>{\centering\arraybackslash}X}
\newcolumntype{L}{>{\raggedright\arraybackslash}X}
\newcolumntype{R}{>{\raggedleft\arraybackslash}X}
\begin{tabularx}{\textwidth}{LLCCC}
\toprule
\multicolumn{2}{c}{} & Pass@1 & Pass@3 & Pass@5 \\
\midrule
\multicolumn{2}{l}{Human} & 0.333 & 0.661 & 0.778 \\
\midrule
\multicolumn{2}{l}{DQN-expert (phyre)} & 0.000 & 0.056 & 0.056 \\
\midrule
\multicolumn{5}{l}{\textit{Open-Sourced Baselines}} \\
\midrule
\multicolumn{2}{l}{Qwen3-VL-8B-Instruct} & 0.056 & 0.167 & 0.167 \\
\multicolumn{2}{l}{Qwen3-VL-8B (SFT)} & 0.0 & 0.167 & 0.222 \\
\midrule
\multicolumn{5}{l}{\textit{Closed-Source API}} \\
\midrule
\multicolumn{2}{l}{Gemini-3.1-Flash} & \underline{0.111} & \textbf{0.278} & 0.278 \\
\multicolumn{2}{l}{Gemini-3.1-Pro} & \underline{0.111} & \textbf{0.278} & \textbf{0.389} \\
\midrule
\multicolumn{5}{l}{\textit{\textbf{Ours (base model: Qwen3VL-8B-SFT)}}} \\
\midrule
\multicolumn{2}{l}{+EG+VAORA} & \textbf{0.167} & \underline{0.222} & \underline{0.333} \\
\bottomrule
\end{tabularx}
\caption{\textbf{Results on Virtual-Tool.} Pass@$k$ results on Virtual-Tool, evaluating cross-engine generalization of models trained solely on Phyre. \ours achieves the best Pass@1 and remains competitive with frontier closed-source VLMs on Pass@3 and Pass@5, despite never being exposed to Virtual-Tool during training.}
\label{tab:virtual_tool}
\vspace{-3mm}
\end{table*}
To assess our generalization, we directly deploy the PHYRE-trained model on Virtual Tool, an entirely different physics simulation environment, without any additional fine-tuning or environment-specific adaptation.
As shown in~\cref{tab:virtual_tool}, \ours matches or outperforms Gemini-3.1-Flash across all metrics and remains competitive with Gemini-3.1-Pro, despite being a compact 8B model trained exclusively on PHYRE with no exposure to Virtual Tool. In sharp contrast, the DQN agent, a strong in-domain expert on PHYRE, achieves near-zero performance on Virtual Tool, highlighting the fundamental inability of non-VLM agents to transfer beyond their training distribution. These results confirm that \ours develops physical understanding that is genuinely environment-agnostic, rather than a collection of PHYRE-specific visual shortcuts.

\subsection{Broader Physical Understanding on CRAFT VQA}
\begin{table*}[t]
\footnotesize 
\centering
\setlength{\tabcolsep}{4pt} 
\newcolumntype{C}{>{\centering\arraybackslash}X}

\begin{tabularx}{\textwidth}{lCCCCCC}
\toprule
Method & Descriptive & Counterfactual & Enable & Causal & Prevent & \textbf{Overall} \\
\midrule
Qwen3-VL-8B-Instruct & 26.60 & 41.70 & 40.70 & 42.00 & 40.70 & 38.34 \\
SFT                 & 37.70 & 46.90 & 43.90 & 44.70 & 47.50 & 44.14 \\
+EG              & 38.50 & 44.50 & 46.00 & 42.30 & \textbf{51.50} & 44.56 \\
+EG+VAORA        & \textbf{39.40} & \textbf{48.20} & \textbf{46.30} & \textbf{45.90} & 50.50 & \textbf{46.06} \\
\bottomrule
\end{tabularx}
\caption{\textbf{CRAFT VQA Benchmark Performance.} Training exclusively on PHYRE, \ours achieves consistent improvement across all CRAFT VQA question types, validating that the learned physical reasoning transfers beyond the interactive training environment.}
\label{tab:CRAFT_VQA}
\vspace{-6mm}
\end{table*}
Beyond task execution, we evaluate whether \ours improves the model's general
physical understanding on CRAFT VQA, which probes five categories of physical
reasoning: descriptive, counterfactual, enable causal, prevent causal, and
overall accuracy. \Cref{tab:CRAFT_VQA} shows the results.
As shown in~\cref{tab:CRAFT_VQA}, \ours consistently improves over the SFT, EG, and the base model across all five reasoning categories, demonstrating that the physical understanding developed through PHYRE training is not confined to action grounding but extends to broader causal reasoning. Notably, the most pronounced gains appear in the counterfactual and causal categories, precisely the reasoning types that require understanding of interactive physical consequences rather than static visual descriptions. This suggests that training a model to reason about the outcomes of its own actions naturally induces a deeper causal understanding of the physical world.

\subsection{The Necessary Condition: Interactive Outcome Alignment}
\begin{table*}[t]
\footnotesize 
\centering
\setlength{\tabcolsep}{3pt} 
\newcolumntype{C}{>{\centering\arraybackslash}X}

\begin{tabularx}{\textwidth}{lCCCCCCCCC} 
\toprule
& \multicolumn{3}{c}{Testing Set 1} & \multicolumn{3}{c}{Testing Set 2} & \multicolumn{3}{c}{Testing Set 3} \\
\cmidrule(lr){2-4} \cmidrule(lr){5-7} \cmidrule(lr){8-10}
Method & Grounding & Placement & Collision & Grounding & Placement & Collision & Grounding & Placement & Collision \\
\midrule
Qwen3-VL-8B-Instruct & 0.017 & 0.022 & 0.0 & 0.013 & 0.029 & 0.0 & 0.015 & 0.023 & 0.0 \\
SFT & 0.710 & 0.348 & 0.255 & 0.670 & 0.496 & 0.299 & 0.695 & 0.297 & 0.419 \\
+EG & 0.424 & 0.0 & 0.401 & 0.639 & 0.512 & 0.297 & 0.686 & 0.523 & 0.373 \\
+EG+VAORA (\ours) & \textbf{0.753} & \textbf{0.586} & \textbf{0.511} & \textbf{0.699} & \textbf{0.641} & \textbf{0.407}  & \textbf{0.739} & \textbf{0.640} & \textbf{0.525} \\
\bottomrule
\end{tabularx}
\caption{\textbf{Reward Breakdown across Cross-Validation Sets.} Supervised finetuning improves only non-interactive grounding, EG risks reward collapse, and \ours consistently achieves the highest grounding, placement, and collision rewards across all testing sets.}
\label{tab:alignment_reward}
\end{table*}
To understand \textit{why} models succeed or fail on unseen tasks, we analyze the reward breakdown across all three testing configurations in~\cref{tab:alignment_reward}, measuring whether the model's chain-of-thought reasoning --- grounding, placement, and collision --- truly generalizes to held-out tasks.
As shown in~\cref{tab:alignment_reward}, SFT successfully transfers the ability to ground static scene perception, a visual alignment capability that does not require interaction, but consistently fails on placement and collision rewards across all testing sets, confirming that imitating reasoning traces provides no supervision over the physical consequences that unfold after the agent acts. With only the Expert-Guided success-probability reward, the model abandons grounded reasoning in favor of shortcut visual-to-action mappings: grounding degrades significantly, and placement collapses entirely on testing set 1, decoupling action quality from any meaningful chain-of-thought.

+EG+VAORA not only recovers the grounding quality lost under shortcut learning but substantially improves placement and collision rewards across all three testing sets. This confirms that VAORA provides the critical missing signal that aligns the model's reasoning with the actual physical consequences of its actions, directly explaining the consistent performance gains of +EG+VAORA across all benchmarks in~\cref{tab:phyre_main},~\cref{tab:virtual_tool}, and~\cref{tab:CRAFT_VQA}.

\subsection{Decomposing VAORA: The Contribution of Each Reward Component}
\begin{table}[t]
\small
\centering
\newcolumntype{C}{>{\centering\arraybackslash}X}
\newcolumntype{L}{>{\raggedright\arraybackslash}X}
\begin{tabularx}{\columnwidth}{LLCCC}
\toprule
\multicolumn{2}{l}{Method} & Pass@1 & Pass@3 & Pass@5 \\
\midrule
\multicolumn{2}{l}{SFT} & 0.176 & 0.368 & 0.452 \\
\multicolumn{2}{l}{+EG} & 0.218 & 0.340 & 0.392 \\
\multicolumn{2}{l}{+EG+VAORA($r_G$)} & 0.272 & 0.432 & 0.488 \\
\multicolumn{2}{l}{+EG+VAORA($r_G$, $r_P$)} & 0.286 & 0.420 & 0.486 \\
\multicolumn{2}{l}{+EG+VAORA($r_G$, $r_C$)}& 0.282 & 0.426 & 0.476 \\
\multicolumn{2}{l}{+EG+VAORA($r_G$, $r_P$, $r_C$)} & 0.366 & 0.474 & 0.514 \\
\bottomrule
\end{tabularx}
\caption{\textbf{Ablation Study of Alignment Components} We demonstrate that supervising a broader set of environmental outcomes consistently yields higher generalization capability.}
\label{tab:ablation_study}
\vspace{-6mm}
\end{table}

We conduct an ablation study on Testing Set~3 to examine the individual contribution of each VAORA reward component: grounding ($r_G$), placement 
($r_P$), and collision ($r_C$). As shown in~\cref{tab:ablation_study}, each component provides meaningful and complementary gains.
\section{Conclusion}
\label{sec:conclusion}
We identified two fundamental failure modes limiting VLM generalization in interactive physical reasoning, hallucinated chain-of-thought and reasoning-action misalignment, and introduced VAORA, an outcome-aligned reward design that directly supervises the alignment between the model's reasoning and the visual outcome of its actions, stabilized by Expert-Guided dense rewards from a pretrained DQN expert. Across cross-task generalization, cross-environment transfer, and physical reasoning VQA, VAORA consistently surpasses task-specific RL experts and matches or outperforms frontier closed-source models, demonstrating that grounded reasoning is the key to generalizable physical intelligence.

\paragraph{Limitation}
Despite these gains, our approach operates in a single-turn setting, committing to a single action prediction without observing the outcome of prior attempts. This is a fundamental constraint on generalization: when physical concepts at test time differ substantially from those seen during training, or when the dynamics of the target environment deviate significantly (e.g., large differences in gravitational constants or friction coefficients), the model has no mechanism to detect and recover from its errors.

\paragraph{Future Work}
Closing this gap points to a clear direction for future work: \textit{test-time adaptive reasoning} and \textit{multi-turn interaction}, where the model updates its physical understanding in response to environmental feedback across successive attempts, mirroring the trial-and-error process that underlies human physical problem-solving.

\newpage
\renewcommand{\bibname}{References}
{
\small
\bibliographystyle{abbrvnat}
\bibliography{references} 
}

\newpage
\appendix


\appendix

\section{Benchmark Details and Adaptations}
\label{app:benchmarks}

\subsection{PHYRE}
\label{app:phyre}

PHYRE~\citep{bakhtin2019phyre} is a physics simulation benchmark requiring an agent to 
place a single red ball with an adjustable radius to achieve a goal state (\emph{e.g.}, 
making green objects touch the blue object when the simulation starts). The action space 
is continuous, $\mathbf{a} = [x, y, r] \in \mathbb{R}^3$, and task success is binary and 
sparse, making it an ideal testbed for our training framework. The benchmark comprises 25 
task types, each containing 100 tasks. For cross-task generalization experiments, we follow 
a held-out protocol: 20 task types (2,000 tasks) are used for training and 5 task types 
(500 tasks) are held out as the testing set. Neither the SFT stage nor the \ours training 
stage exposes any tasks from the held-out testing types, ensuring that all reported gains 
reflect genuine generalization to unseen task distributions. For cross-environment and 
broader physical understanding experiments, the model is trained on all 25 task types 
(2,500 tasks).

\subsection{Virtual Tool}
\label{app:virtual_tool}

Virtual Tool is a distinct physics simulation environment involving tool-use reasoning. We evaluate on the Original Split of the Virtual Tool benchmark, which contains the most basic tool-use tasks. To enable direct evaluation of models trained on PHYRE without any environment-specific re-training, we modify the task protocol in two ways: we remove the tool selection step and instead allow models to attempt each tool multiple times per task, and we remap the colors of the target zone and the object to match the analogous colors used in PHYRE tasks, improving the visual adaptability of models trained on PHYRE. A task is considered solved at the minimum number of attempts required across all tools, and we report Pass@k under this definition.

\subsection{CRAFT VQA}
\label{app:craft_vqa}

CRAFT VQA is a diagnostic benchmark for physical understanding that probes 
descriptive, counterfactual, and causal reasoning through visual question answering. Since 
the original CRAFT VQA dataset takes full simulation videos as input, we adapt the 
benchmark to use only the initial scene image of each simulation, aligning with our 
single-image training pipeline. For evaluation, we randomly sample 1,000 questions from 
each question type defined by CRAFT VQA.

\section{Implementation Details}
\label{app:implementation}

\subsection{SFT Dataset Construction}
\label{app:sft}

To construct the SFT dataset, we randomly sample up to 20 successful actions per task across all 2,500 tasks, spanning all task types; if fewer than 20 successful actions are available for a given task, we retain them all. Each sampled action is executed in the simulator, and grounding, placement, and collision information are extracted from the resulting simulation. For each action and its associated simulation information, we prompt Gemini-3.1-Flash-Preview to generate two distinct structured reasoning traces conditioned on this information, yielding a final dataset of 47,177 reasoning traces in total. For the cross-task generalization experiments, we partition the SFT dataset such that tasks appearing in the test split are strictly excluded from SFT training, ensuring no leakage between training and evaluation. \ours is subsequently trained by jointly optimizing gated visual-action alignment rewardd and visual alignment reward under GDPO on top of this SFT base.

\subsection{PHYRE Training and Testing Split}
\label{app:phyre_split}

\begin{figure}[t]
    \centering
    \includegraphics[width=\linewidth]{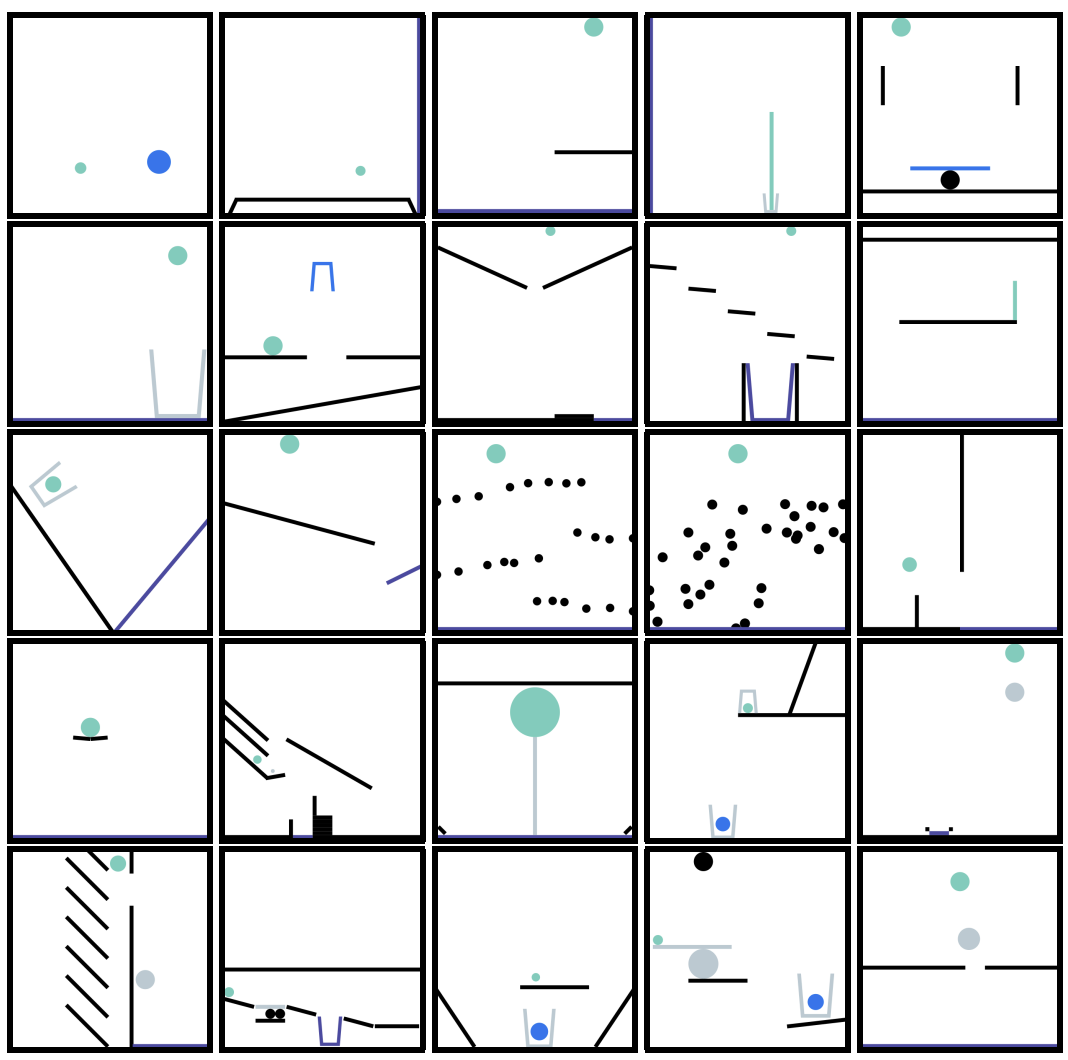}
    \caption{Overview of all 25 task types in the PHYRE benchmark, 
    arranged in row-major order from task type 00000 (top-left) to 00024 
    (bottom-right).~\citep{bakhtin2019phyre}.}
    \label{fig:phyre_task_types}
\end{figure}

The PHYRE benchmark comprises 25 task types, each containing 100 tasks. Figure~\ref{fig:phyre_task_types} illustrates all 25 task types, where the top-left corresponds to task type \texttt{00000} and the bottom-right to \texttt{00024}. For the cross-task generalization experiments, we adopt a held-out protocol: 20 task types (2,000 tasks) are used for training and 5 task types (500 tasks) are held out as the testing set. Neither the SFT stage nor the \ours training stage exposes any tasks from the held-out testing types, ensuring that all reported gains reflect genuine generalization to unseen task distributions.
\paragraph{Selection Criteria for Testing Task Types.} Since our collision-interactive reward is designed to model only the \emph{first} object that the red ball collides with upon placement, the generalization capability we aim to evaluate is meaningful only for tasks in which the key interaction is itself the first collision event triggered by the red ball — that is, tasks that do not require long action chains in which multiple intermediate collisions must occur before the critical interaction. Based on this criterion, 10 out of the 25 task types are suitable candidates for the held-out testing set: 00000, 00001, 00002, 00003, 00006, 00009, 00012, 00013, 00014, and 00015. Inspecting Figure~\ref{fig:phyre_task_types}, these tasks share the property that the placed object interacts directly and immediately with the target or a single intermediary, without requiring a cascading chain of collisions to achieve the goal.
\paragraph{Three-Split Protocol.} We construct three different held-out splits drawn from the 10 eligible task types and train an independent model on each, reporting results on the corresponding held-out testing set in each case:
\begin{itemize}
\item \textbf{Testing set 1:} 00001, 00002, 00003, 00006, 00013
\item \textbf{Testing set 2:} 00000, 00001, 00003, 00012, 00014
\item \textbf{Testing set 3:} 00000, 00001, 00002, 00009, 00015
\end{itemize}
This three-split protocol verifies that the performance gains of \ours are stable across different choices of held-out task types and are not an artifact of a particular train/test partition. For cross-environment generalization on Virtual Tool and broader physical understanding on CRAFT VQA, the model is trained on all 25 task types (2,500 tasks) without any held-out split.

\subsection{Training of the in-domain DQN Expert}
\label{app:dqn}
The DQN expert follows the training procedure described in the original PHYRE paper~\cite{bakhtin2019phyre}, and is trained exclusively on the PHYRE-B tier using the offline DQN variant.
The training data used to fit the expert depends on the experimental setting.
In the cross-task experiments, the expert is trained only on tasks belonging to the training set, ensuring no leakage of information about held-out evaluation tasks.
In the full-task experiments, where the model is evaluated on Virtual Tool and CraftVQA, the expert is trained on all available tasks in PHYRE.
In both cases, the expert is kept frozen after training and serves solely as a success probability estimator to gate the VAORA reward signal.

\subsection{VAORA: Visual Action Outcome Reasoning Alignment}
\label{app:vaora}
\paragraph{Textual Spatial Label}
Textual spatial labels provide a discrete, structured vocabulary for expressing spatial relationships in natural language.
All labels are drawn from a fixed set of nine regions arranged in a $3\times3$ grid:
\{\texttt{TOP-LEFT}, \texttt{TOP}, \texttt{TOP-RIGHT}, \texttt{LEFT}, \texttt{CENTER}, \texttt{RIGHT}, \texttt{BOTTOM-LEFT}, \texttt{BOTTOM}, \texttt{BOTTOM-RIGHT}\}.
The interpretation of these regions differs between the grounding reward and the placement reward, as described below.

\begin{figure}[ht]
    \centering
    \includegraphics[width=0.45\textwidth]{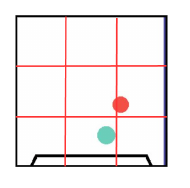}
    \caption{The $3\times3$ spatial grid defined over the full scene for the grounding reward.
    Each region corresponds to one of the nine textual spatial labels.}
    \label{fig:grid_full}
\end{figure}

For scene \textbf grounding, the $3\times3$ grid is defined over the \emph{entire scene}, dividing the full $256\times256$ image into nine equal regions.
Each object's label $\hat{\ell}^{(i)}$ is assigned by mapping its predicted coordinates to the corresponding region.
Figure~\ref{fig:grid_full} illustrates this grid overlaid on an example scene.

\begin{figure}[ht]
    \centering
    \includegraphics[width=0.45\textwidth]{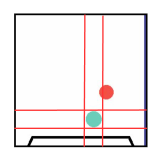}
    \caption{The $3\times3$ spatial grid defined relative to the green ball (highlighted in green) for the placement reward.
    Each region describes the spatial relationship of the placed ball with respect to the reference object.}
    \label{fig:grid_obj}
\end{figure}

For placement grounding, the $3\times3$ grid is defined \emph{relative to a reference object} $\widehat{\text{obj}}^{(i)}$, centered on that object with each cell sized proportionally to the object's bounding box.
The label $\hat{\ell}^{(i)}$ thus describes where the placed ball is located relative to the reference object (e.g., \texttt{TOP-RIGHT} of the green ball), rather than its absolute position in the scene.
Figure~\ref{fig:grid_obj} illustrates this object-relative grid centered on the green ball.

\paragraph{Soft-Grid Score.}
The soft-grid score computes a continuous proximity measure between two spatial labels by embedding them into a $3\times3$ grid.
Each canonical region label is mapped to a grid coordinate $(c_x, c_y) \in \{0,1,2\}^2$, corresponding to its column and row position in the grid (e.g., \texttt{TOP-LEFT}$\mapsto(0,0)$, \texttt{CENTER}$\mapsto(1,1)$, \texttt{BOTTOM-RIGHT}$\mapsto(2,2)$).
Given a predicted label $\hat{\ell}$ with grid coordinate $(\hat{c}_x, \hat{c}_y)$ and a ground-truth label $\ell$ with grid coordinate $(c_x, c_y)$, the soft-grid score is defined as:

\begin{equation}
    \mathrm{soft}(\hat{\ell},\,\ell) = 1 - \frac{\sqrt{(\hat{c}_x - c_x)^2 + (\hat{c}_y - c_y)^2}}{2\sqrt{2}}
\end{equation}

\noindent The denominator $2\sqrt{2}$ corresponds to the maximum possible distance in the $3\times3$ grid (i.e., between diagonally opposite corners), normalizing the score to $[0, 1]$.
A score of $1$ indicates a perfectly matched label, while $0$ indicates maximally distant labels, allowing the reward to penalize spatial misalignment gracefully rather than with a hard binary criterion.

\paragraph{Hyperparameter Setting: Weights and Penalties.}
We set the reward weights as $w_g = 0.2$, $w_c = 0.6$, and $w_p = 0.3$, reflecting the relative importance of each reward component and remaining adjustable to different environment settings.
The per-item penalty coefficients are set to $p_g = \tfrac{1}{2}$, $p_c = \tfrac{1}{6}$, and $p_p = \tfrac{1}{3}$, and are derived from the constraint $w_{\cdot} \cdot p_{\cdot} = 0.1$, i.e.:
\begin{equation}
    w_g \cdot p_g = 0.2 \times \tfrac{1}{2} = 0.1, \quad
    w_c \cdot p_c = 0.6 \times \tfrac{1}{6} = 0.1, \quad
    w_p \cdot p_p = 0.3 \times \tfrac{1}{3} = 0.1.
\end{equation}
This design ensures that formatting errors and hallucinated outputs are penalized uniformly at $-0.1$ per item across all reward types, providing a consistent and reward-scale-invariant signal for discouraging malformed or inaccurate reasoning regardless of which component detects the violation.

\subsection{Training Configurations}
\label{app:train_config}

All experiments are conducted on two NVIDIA H200 GPUs.

\paragraph{Cross-Task Training.}
We evaluate generalization under a cross-task protocol, where models are trained on three different subsets of tasks of the PHYRE environment and evaluated on held-out tasks unseen during training.
All cross-task experiments use Qwen3-VL-8B as the base model, first supervised fine-tuned on 36,000 reasoning traces collected exclusively from the training-set tasks (ensuring no leakage from held-out evaluation tasks) for 3 epochs using LoRA with rank $r=64$ and $\alpha=128$, then trained with the GDPO objective with a learning rate of $1\times10^{-6}$, a rollout sample size of $n=5$, and a maximum response length of $1280$ tokens with $40$ timesteps and choose the best checkpoints based on the training curve.
 
\paragraph{Full-Task Training.}
We further evaluate cross-environment generalization and CraftVQA under a full-task protocol, where the model is trained on all available tasks of the PHYRE environment.
We compare two settings: training \textit{with} VAORA rewards and training \textit{without} VAORA rewards (i.e., action reward only), to isolate the contribution of our reward design.
\textit{(1) Full-task training with VAORA.} \textit{(2) Full-task training without VAORA.}
Both experiments use Qwen3-VL-8B as the base model, first supervised fine-tuned on all 47,177 reasoning traces for 3 epochs using LoRA with rank $r=64$ and $\alpha=128$, then trained with the GDPO objective with a learning rate of $1\times10^{-6}$, a rollout sample size of $n=5$, and a maximum response length of $1280$ tokens with $80$ timesteps, and choose the best checkpoints based on the training curve.

\subsection{Inference Configurations}
\label{app:inf_config}
All experiments are conducted on one NVIDIA H200 GPU.

\paragraph{Interactive Physical Reasoning Tasks.}
For inference on interactive physical reasoning tasks, namely PHYRE and Virtual Tool, we use a sampling temperature of $0.7$ and top-$p$ of $0.9$ for all baselines.
These settings encourage sufficient output diversity to account for the stochastic nature of physical interaction, while maintaining coherent and grounded reasoning traces.

\paragraph{Visual Question Answering Tasks.}
For inference on VQA tasks, including CraftVQA, we adopt a lower sampling temperature of $0.2$ and top-$p$ of $0.95$ for all baselines.
The reduced temperature reflects the more deterministic nature of question answering, where precise and consistent outputs are preferred over diverse sampling.

\subsection{Further Details and Codebase Release}
Further details are released at~\url{https://vaora-proj.github.io/}, including datasets, checkpoints, and the codebase.

\section{Qualitative Results}
\subsection{PHYRE: Reasoning Trace}

\begin{figure}[ht]
    \centering
    \begin{subfigure}[b]{0.32\textwidth}
        \includegraphics[width=\linewidth]{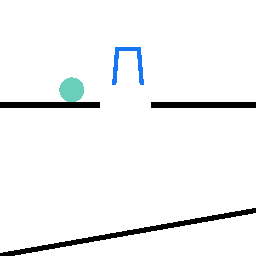}
        \caption*{Initial Scene}
    \end{subfigure}
    \hfill
    \begin{subfigure}[b]{0.32\textwidth}
        \includegraphics[width=\linewidth]{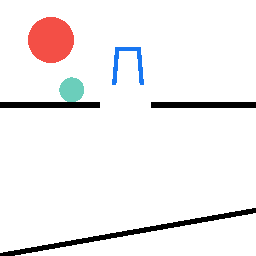}
        \caption*{Frame 0}
    \end{subfigure}
    \hfill
    \begin{subfigure}[b]{0.32\textwidth}
        \includegraphics[width=\linewidth]{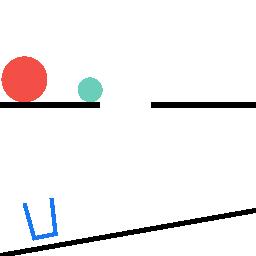}
        \caption*{Frame 1}
    \end{subfigure}

    \vspace{0.5em}

    \begin{subfigure}[b]{0.32\textwidth}
        \includegraphics[width=\linewidth]{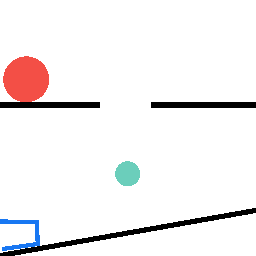}
        \caption*{Frame 2}
    \end{subfigure}
    \hfill
    \begin{subfigure}[b]{0.32\textwidth}
        \includegraphics[width=\linewidth]{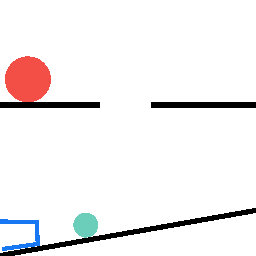}
        \caption*{Frame 3}
    \end{subfigure}
    \hfill
    \begin{subfigure}[b]{0.32\textwidth}
        \includegraphics[width=\linewidth]{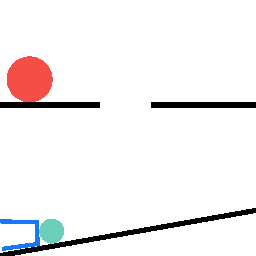}
        \caption*{Frame 4}
    \end{subfigure}

    \caption{
        \textbf{PHYRE qualitative example.}
        The initial scene (top-left) shows a \textcolor{green}{green ball} resting on a left platform and a \textcolor{blue}{blue jar} positioned to its right.
        The model correctly identifies the scene configuration, reasons that placing the \textcolor{red}{red ball} at the upper-left of the green ball will strike it rightward, and predicts the action $[51, 39, 23]$.
        The simulation frames confirm that the reasoning trace is action-aligned: the red ball falls, strikes the green ball, which subsequently slides off the platform and enters the blue jar — achieving the goal.
    }
    \label{fig:qualitative_phyre}
\end{figure}

We present a qualitative example demonstrating how our model produces a reasoning trace that is both physically grounded and action-aligned in Phyre cross-task configuration.
Figure~\ref{fig:qualitative_phyre} shows the initial scene alongside five consecutive simulation frames following the model's predicted action.

\noindent The reasoning trace produced by our model illustrates two key properties.
First, the \texttt{<scene\_answer>} section accurately localizes all objects in the initial scene — the green ball on the left platform at approximately $(71, 94)$ and the blue jar above at $(129, 70)$ — demonstrating faithful scene grounding with no hallucination.
Second, the \texttt{<causal\_actions\_answer>} section specifies a physically correct interaction: striking the green ball on its upper-left side at $(50, 49)$ to impart rightward momentum, which is precisely what the simulation confirms across Frames~0--4.
This alignment between the reasoning trace and the observed visual outcome validates that VAORA successfully suppresses both hallucinated CoT and misaligned action, producing predictions that are grounded, causal, and consistent with physical reality.

\noindent The full reasoning trace generated by our model for this example is shown below:
 
\begin{lstlisting}[basicstyle=\ttfamily\scriptsize, breaklines=true, frame=single, label={lst:phyre_trace}, caption={Full reasoning trace for the PHYRE qualitative example.}]
<scene_reasoning>
The scene consists of a green ball resting on a horizontal black platform on the left
side. To the right of the green ball, there is a blue jar standing upright. A second
horizontal black platform is positioned below and to the right of the blue jar,
separated from the left platform by a gap. The green ball is contained within the
boundaries of the left platform. To solve the puzzle, the green ball must be moved
from its current position, through the gap, and into the blue jar. Since the green
ball is stationary, an external force is required to initiate movement towards the right.
</scene_reasoning>
 
<scene_answer>
[BLACK BAR] is at [LEFT] [49, 108] with size [128]
[GREEN BALL] is at [LEFT] [71, 94] with size [24]
[BLUE JAR] is at [TOP] [129, 70] with size [48]
</scene_answer>
 
<causal_actions_reasoning>
To move the green ball into the blue jar, it needs to gain horizontal momentum directed
toward the right. By striking the green ball on its upper-left side, the impact will
transfer kinetic energy and provide a downward-and-rightward force. Specifically,
hitting the ball at [50, 49] ensures the vector of the force is angled correctly to
overcome the ball's inertia and push it off the left platform toward the opening where
the blue jar is located.
</causal_actions_reasoning>
 
<causal_actions_answer>
The [RED BALL] should [STRIKE] the [GREEN BALL] at [50, 49] to push it towards [RIGHT].
</causal_actions_answer>
 
<placement_reasoning>
To achieve the necessary impact at [50, 49], the red ball must be placed at a position
that allows gravity to accelerate it into the target contact point. Placing the red ball
at [51, 39] aligns it slightly to the left and above the green ball. Given the ball's
radius of 23, it will fall vertically and strike the green ball on its left curve. This
specific placement ensures the red ball has enough clearance to fall and that the
resulting collision provides the rightward trajectory needed for the green ball to reach
the blue jar.
</placement_reasoning>
 
<placement_answer>
[RED BALL] is located at the [TOP-LEFT] of the [WHOLE SCENE].
[RED BALL] is located at the [TOP-LEFT] of the [GREEN BALL].
</placement_answer>
 
<action>
[51, 39, 23]
</action>
\end{lstlisting}

\subsection{Virtual-Tool: Reasoning Trace}
 
\begin{figure}[ht]
    \centering
    \begin{subfigure}[b]{0.32\textwidth}
        \includegraphics[width=\linewidth]{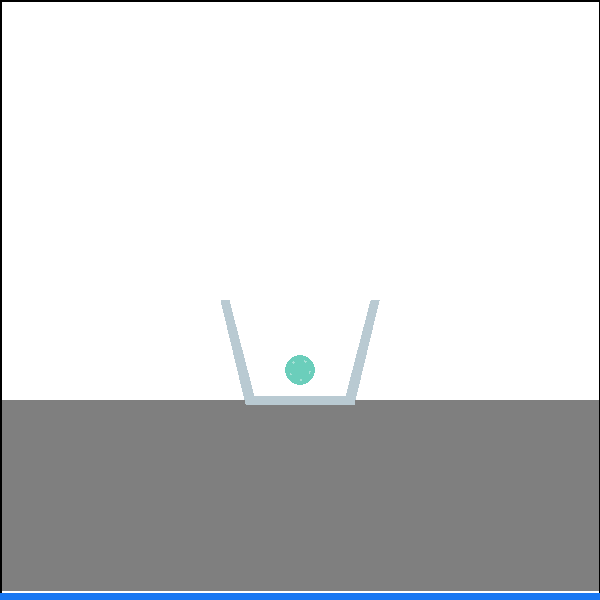}
        \caption*{Initial Scene}
    \end{subfigure}
    \hfill
    \begin{subfigure}[b]{0.32\textwidth}
        \includegraphics[width=\linewidth]{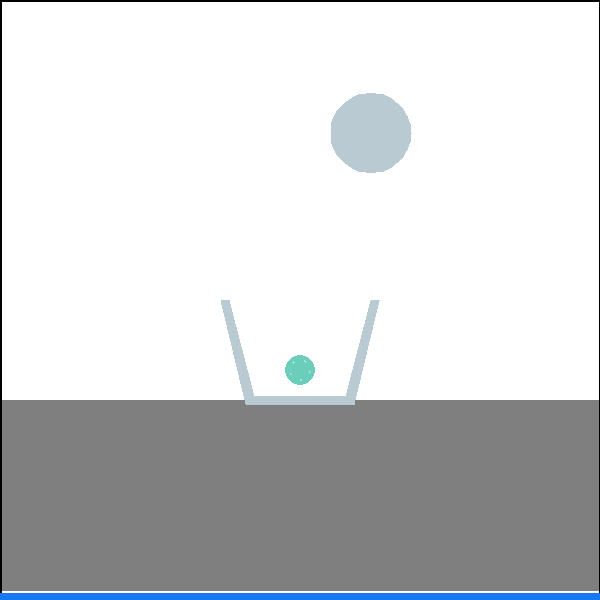}
        \caption*{Frame 0}
    \end{subfigure}
    \hfill
    \begin{subfigure}[b]{0.32\textwidth}
        \includegraphics[width=\linewidth]{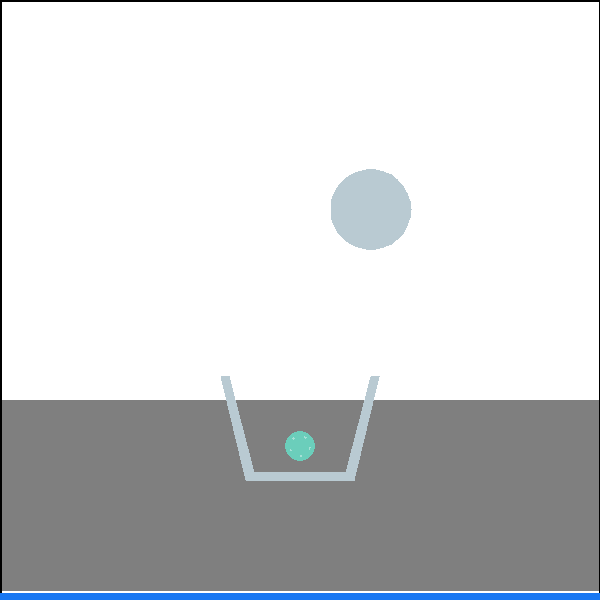}
        \caption*{Frame 1}
    \end{subfigure}
 
    \vspace{0.5em}
 
    \begin{subfigure}[b]{0.32\textwidth}
        \includegraphics[width=\linewidth]{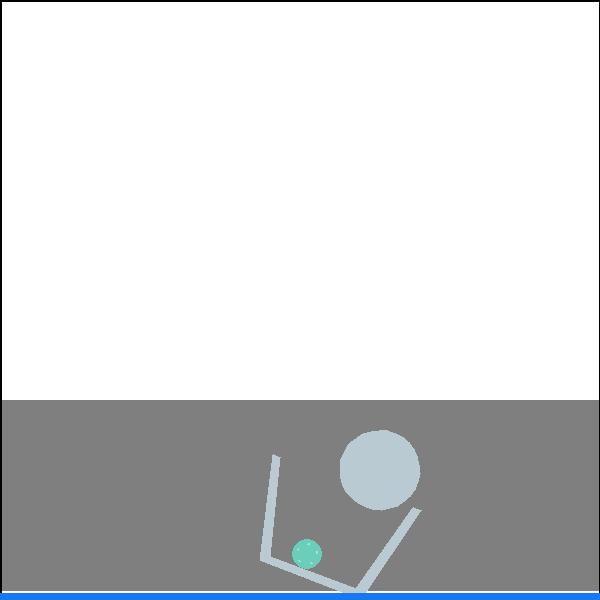}
        \caption*{Frame 2}
    \end{subfigure}
    \hfill
    \begin{subfigure}[b]{0.32\textwidth}
        \includegraphics[width=\linewidth]{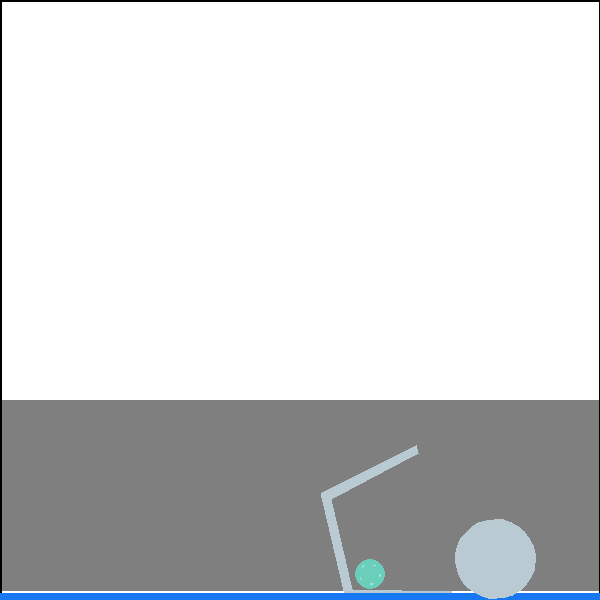}
        \caption*{Frame 3}
    \end{subfigure}
    \hfill
    \begin{subfigure}[b]{0.32\textwidth}
        \includegraphics[width=\linewidth]{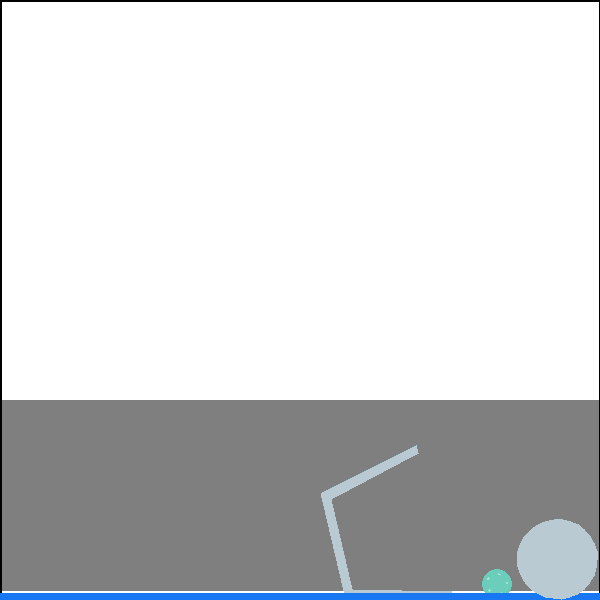}
        \caption*{Frame 4}
    \end{subfigure}
 
    \caption{
        \textbf{Virtual-Tool qualitative example.}
        The initial scene shows a \textcolor{green}{green ball} sitting inside a gray trapezoid container, with a \textcolor{blue}{blue target line} at the bottom of the scene.
        The model correctly reasons to drop a ball onto the trapezoid container to tip it, releasing the green ball toward the target.
        Frames 0--4 confirm the causal chain: the action ball falls and strikes the right side of the trapezoid (Frames 1, 2), causing it to rotate and release the green ball, which then slides along the floor to reach the blue target (Frame 4) — achieving the goal.
        Note that the action object appears as a \emph{gray} ball in the simulation, whereas the model refers to it as a \texttt{[RED OBJECT]} in its reasoning. This minor color discrepancy arises because the model is trained in PHYRE, where the action object is always rendered as a red ball; for consistency, the same \texttt{[RED OBJECT]} terminology is used in the Virtual-Tool prompt. Since the VLM only observes the initial scene — which does not yet contain the action object — this color mismatch does not affect the model's spatial reasoning or placement strategy.
    }
    \label{fig:qualitative_vt}
\end{figure}

We present a qualitative example demonstrating how our model produces a reasoning trace that is both physically grounded and action-aligned in cross-environment Virtual-Tool configuration, which features more complex scene configurations involving a container that must be tipped to release the goal object.
Figure~\ref{fig:qualitative_vt} shows the initial scene alongside five simulation frames following the model's predicted action.
 
\noindent The reasoning trace generated by our model for this example demonstrates strong alignment between scene understanding, causal reasoning, and the observed physical outcome.
The \texttt{<scene\_answer>} section faithfully identifies the green ball sitting inside the gray trapezoid container and the blue target at the bottom — the two objects most critical to solving the task.
The \texttt{<causal\_actions\_answer>} correctly identifies the container as the intermediate object to interact with, reasoning that tilting it will release the green ball downward toward the target, which Frames 2--4 confirm exactly.
The one minor surface-level inconsistency — the action object being described as \texttt{[RED OBJECT]} while rendered as a gray ball in simulation — is a benign artifact of cross-environment prompt reuse and has no effect on the physical validity or action alignment of the reasoning trace.
 
\begin{lstlisting}[basicstyle=\ttfamily\scriptsize, breaklines=true, frame=single, label={lst:virtual-tool_trace}, caption={Full reasoning trace for the Virtual-tool qualitative example.}]
<scene_reasoning>
The scene consists of a green ball resting inside a gray jar, which is positioned on a large gray platform. Below this platform is a blue bar at the bottom of the scene. The objective is to make the green ball touch the blue bar. Currently, the green ball is static and elevated; it is contained within the jar, preventing it from falling directly. To solve this, the jar must be tipped or moved so that the green ball can be released and fall toward the target area below. Since the jar is stable on the platform, an external force must be applied to the jar to destabilize it and initiate movement.
</scene_reasoning>

<scene_answer>
[GRAY BAR] is at [Bottom] [128, 205] with size [256]
[GRAY JAR] is at [Center] [128, 165] with size [68]
[GREEN BALL] is at [Center] [133, 161] with size [16]
[BLUE BAR] is at [Bottom] [128, 252] with size [256]
</scene_answer>

<causal_actions_reasoning>
To release the green ball, the gray jar needs to be tilted or pushed sideways. By hitting the top-right rim of the gray jar at coordinate [160, 113], a lateral force is applied that creates a torque, causing the jar to tilt toward the left. As the jar tilts, the green ball loses its stable support and will roll toward the left side of the jar's opening. Once the ball clears the lip of the jar, gravity will pull it downward toward the blue bar. The specific contact point at [160, 113] is optimal because it is high enough on the jar to maximize the tipping moment while ensuring the force is directed to push the jar away from the center.
</causal_actions_reasoning>

<causal_actions_answer>
The [RED OBJECT] should [HIT] the [GRAY JAR] at [160, 113] to push it towards [LEFT].
The chain actions cause the [GREEN BALL] to move [LEFT] at point [133, 161].
</causal_actions_answer>

<placement_reasoning>
To hit the gray jar at the intended contact point of [160, 113], a red object must be placed directly above this target. Placing the red object at x=160 ensures vertical alignment for gravity to pull it straight down onto the right side of the jar. The y-coordinate of 54 is sufficiently high to allow the object to gain enough kinetic energy and momentum through gravitational acceleration before impact. A radius of 25 provides a large enough surface area and mass to ensure a significant collision, effectively transferring enough energy to tip the jar over and release the green ball.
</placement_reasoning>

<placement_answer>
[RED OBJECT] is located at the [Top] of the [WHOLE SCENE].
[RED OBJECT] is located at the [Top-Right] of the [GRAY JAR].
</placement_answer>

<action>
[160, 54, 25]
</action>"
\end{lstlisting}
\subsection{CraftVQA: Reasoning Trace}
 
\begin{figure}[ht]
    \centering
    \includegraphics[width=0.45\textwidth]{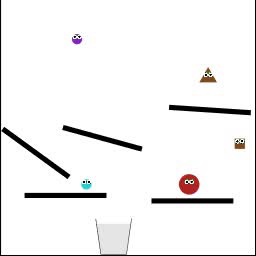}
    \caption{
        \textbf{CraftVQA qualitative example.}
        The scene contains a \textbf{brown triangle} resting on a right-side platform, a gray jar (basket) at the bottom center, and several other dynamic objects (red ball, cyan ball, purple ball, brown ball) on various platforms.
        The question asks: \textit{``If any of the other objects are removed, will the small brown triangle fall into the basket?''}
        The model correctly answers \texttt{true}, reasoning that without other objects blocking or disturbing the trajectory, the brown triangle will fall off the right platform, slide leftward under gravity, and descend directly into the jar.
    }
    \label{fig:qualitative_craft}
\end{figure}

We present a qualitative example demonstrating how our model produces a reasoning trace that is physically grounded and leads to a correct answer on a  physical reasoning question.
Figure~\ref{fig:qualitative_craft} shows the input scene alongside the model's full reasoning trace.
 
\noindent The reasoning trace demonstrates two key properties.
First, the \texttt{<scene\_answer>} section faithfully localizes all nine objects in the scene, correctly placing the \texttt{[BROWN TRIANGLE]} at \texttt{TOP-RIGHT} and the \texttt{[GRAY JAR]} at \texttt{BOTTOM}, which are the two objects central to answering the counterfactual question.
Second, the \texttt{<causal\_actions\_answer>} correctly traces the physical event chain: once other objects are removed, the brown triangle falls off the right platform, moves leftward under gravity, and descends into the jar — leading to the correct final answer \texttt{true}.
This example illustrates that VAORA  generalizes beyond action-prediction tasks to support physical reasoning in VQA settings, producing scene descriptions and causal chains that are both accurate and directly informative for answering the question.
 
\noindent The full reasoning trace generated by our model for this example is shown below:
 
\begin{lstlisting}[basicstyle=\ttfamily\scriptsize, breaklines=true, frame=single, caption={Full reasoning trace for the CraftVQA qualitative example.}]
Question: If any of the other objects are removed, will the small brown triangle fall
into the basket?
 
<scene_reasoning>
The scene consists of a series of platforms and several dynamic objects. A red ball and
a cyan ball are positioned on lower platforms, while a small brown triangle sits atop a
higher platform on the right side. Below these elements is a gray jar (the target
container). The brown triangle is currently stable on its platform. To reach the jar,
the brown triangle must be knocked off its platform so that gravity can pull it downward
toward the target. The spatial arrangement shows that the brown triangle is separated
from the jar by a significant horizontal distance and a vertical gap, requiring an
external force to initiate movement toward the left.
</scene_reasoning>
 
<scene_answer>
[BLACK BAR] is at [LEFT] [33, 161] with size [118]
[CYAN BALL] is at [BOTTOM-LEFT] [92, 198] with size [18]
[GRAY JAR] is at [BOTTOM] [120, 242] with size [64]
[RED BALL] is at [BOTTOM-RIGHT] [192, 189] with size [24]
[BLACK BAR] is at [CENTER] [104, 142] with size [116]
[BLACK BAR] is at [RIGHT] [206, 114] with size [96]
[BROWN TRIANGLE] is at [TOP-RIGHT] [216, 89] with size [26]
[BROWN BALL] is at [RIGHT] [242, 150] with size [16]
[PURPLE BALL] is at [TOP-LEFT] [85, 43] with size [16]
</scene_answer>
 
<causal_actions_reasoning>
To solve the puzzle, the brown triangle must be moved from its stable position on the
right-side platform. If the brown triangle falls off the edge of the black bar at
[206, 114], it will enter a free-fall state. Gravity will then pull it downward toward
the gray jar. The trajectory needs to be directed toward the left so that it clears the
lower platforms and falls directly into the opening of the jar located at the bottom
center. Removing the support or applying a lateral force to the brown triangle is the
necessary causal action to initiate this descent.
</causal_actions_reasoning>
 
<causal_actions_answer>
The brown triangle will move left at point [216, 89].
The brown triangle will enter the jar.
</causal_actions_answer>
 
<final_answer>
true
</final_answer>
\end{lstlisting}

\section{Prompts of PHYRE, Virtual-Tool, and CraftVQA}
\label{app:prompts}

\subsection{PHYRE}
\label{app:prompt-phyre}
\begin{lstlisting}[basicstyle=\ttfamily\scriptsize, breaklines=true, frame=single]
You are a physics reasoning agent in the PHYRE environment.

ENVIRONMENT RULES
- The scene is 256x256 pixels with a white background.
- Objects are colored: black, gray, blue, green, purple.
- Static objects: BLACK and PURPLE are immovable.
- Dynamic objects: GRAY, BLUE, GREEN fall under gravity in 2D.
- Scene boundaries are walls/floor.
- Objects shapes: BALL, BAR, JAR.
- The goal: make any [GREEN *] object touch any [BLUE *] dynamic object OR any [PURPLE *] static object.
- You can add exactly one RED BALL as the action: [x, y, r]
  - x in [0, 255] (0=left edge, 255=right edge)
  - y in [0, 255] (0=top edge, 255=bottom edge)
  - r in [2, 32]

OBJECT NAMING CONVENTION (STRICT)
- When you mention an object, always use: [COLOR SHAPE] in UPPERCASE, e.g. [GREEN BALL], [BLUE BALL], [PURPLE BAR], [GRAY JAR].
- The added object must always be written exactly as: [RED BALL]
- EXCEPTION: In reasoning sections only, you may use lowercase natural language for readability.

INPUT
This is the input scene: <image>

TASK (what to do)
1) Identify the goal pair: which [GREEN *] must touch which [BLUE *] or [PURPLE *].
2) Identify blockers: walls, gaps, covers, containers, slopes, platforms, and black obstacles.
3) Choose the main causal strategy (PUSH/ROTATE/HIT/TILT/BLOCKED/SPIN/COLLIDE WITH/DEFLECTED/DEFLECT/STOPPED/SUPPORT/BLOCK/STOP/STRIKE).
4) Estimate coordinates of relevant objects (approximate is OK but must be plausible).
5) Describe the red ball placement relative to objects AND the whole scene.
6) Output the final action [x, y, r].
7) You are encouraged to propose different reasoning in each section and different strategies to improve diversity.

OUTPUT FORMAT (MUST MATCH EXACTLY)
You MUST output EXACTLY these sections in EXACT order, and NOTHING else:

<scene_reasoning>
...
</scene_reasoning>
<scene_answer>
...
</scene_answer>
<causal_actions_reasoning>
...
</causal_actions_reasoning>
<causal_actions_answer>
...
</causal_actions_answer>
<placement_reasoning>
...
</placement_reasoning>
<placement_answer>
...
</placement_answer>
<action>
[x, y, r]
</action>

CONTENT RULES PER SECTION

<scene_reasoning>
REASONING REQUIREMENTS:
- Identify all relevant objects and estimate their approximate positions.
- State the goal clearly (which green touches which blue/purple).
- Analyze spatial relationships: alignment, distance, and location relative to the scene center.
- Identify obstacles between goal objects.
- Assess whether objects will naturally interact or need intervention.
</scene_reasoning>

<scene_answer>
- List the key goal objects (at least the [GREEN *], [GRAY *] and its goal [BLUE *]/[PURPLE *]).
- Include any critical blockers or intermediate objects.
- Use THIS exact line style:
  [OBJECT] is at [GLOBAL SPATIAL] [x, y] with size [d].
- GLOBAL SPATIAL must be one of:
  ['TOP-LEFT', 'TOP-RIGHT', 'CENTER', 'TOP', 'BOTTOM-RIGHT', 'BOTTOM-LEFT', 'LEFT', 'BOTTOM', 'RIGHT']
- x, y, d must be integers in range 0-255.
</scene_answer>

<causal_actions_reasoning>
REASONING REQUIREMENTS:
- Reference coordinates from <scene_answer> when describing positions.
- Choose and justify ONE causal strategy.
- Describe the event chain: red ball -> interaction -> target motion -> goal achievement.
- Explain the contact point and why it creates the desired motion.
- Verify the approach avoids obstacles.
</causal_actions_reasoning>

<causal_actions_answer>
- Output 1-2 lines total using one of the following templates:

Template Block (for * BALL or * JAR):
  ACTION in ['DEFLECT', 'BLOCK', 'STOP']
  The [RED BALL] should [ACTION] the [TARGET OBJECT] at the contact point [x, y].

Template Collision (for * BALL or * JAR):
  ACTION in ['DEFLECTED', 'COLLIDE WITH', 'BLOCKED', 'STOPPED', 'PUSH', 'STRIKE', 'HIT']
  DIRECTION in ['RIGHT', 'LEFT']
  The [RED BALL] should [ACTION] the [TARGET OBJECT] at the contact point [x, y] to push it towards [DIRECTION].

Template Rotation (for * BAR):
  ACTION in ['TILT', 'ROTATE', 'SPIN']
  DIRECTION in ['CLOCKWISE', 'COUNTER-CLOCKWISE']
  The [RED BALL] should [ACTION] the [TARGET OBJECT] at the contact point [x, y] to rotate it [DIRECTION].

Template Support (for * BAR):
  The [RED BALL] should [SUPPORT] the [TARGET OBJECT] at the contact point [x, y].

For chain actions on the GREEN object:
  Template Collision: The chain actions cause the [GREEN BALL or GREEN JAR] to move [LEFT|RIGHT] at point [x, y].
  Template Block:     The chain actions cause the [GREEN BALL or GREEN JAR] to be [STOPPED|DEFLECTED|BLOCKED] at point [x, y].
  Template Rotation:  The chain actions cause the [GREEN BAR] to move [CLOCKWISE|COUNTER-CLOCKWISE] at point [x, y].
</causal_actions_answer>

<placement_reasoning>
REASONING REQUIREMENTS:
- Reference target position from <scene_answer> and contact point from <causal_actions_answer>.
- Calculate red ball initial position to achieve the desired contact.
- Account for ball radius and gravity trajectory.
- Justify radius choice (larger for momentum, smaller for precision).
- CRITICAL PHYSICS RULE: To push an object to a [DIRECTION], the [RED BALL] must be placed on the OPPOSITE side.
</placement_reasoning>

<placement_answer>
- Output 1-3 lines using these templates:
  [RED BALL] is located at the [RELATIVE SPATIAL] of the [TARGET OBJECT A].
  [RED BALL] is located at the [RELATIVE SPATIAL] of the [TARGET OBJECT B].
  [RED BALL] is located at the [GLOBAL SPATIAL] of the [WHOLE SCENE].

- RELATIVE SPATIAL in ['TOP-LEFT', 'TOP-RIGHT', 'BOTTOM-LEFT', 'BOTTOM-RIGHT']
- GLOBAL SPATIAL in ['TOP-LEFT', 'TOP-RIGHT', 'CENTER', 'TOP', 'BOTTOM-RIGHT', 'BOTTOM-LEFT', 'LEFT', 'BOTTOM', 'RIGHT']
</placement_answer>

<action>
- Output ONLY: [x, y, r]
- x, y in [0, 255], r in [2, 32]
- Do not add any explanation here.
</action>
\end{lstlisting}

\subsection{Virtual-Tool}
\label{prompt-virtual-tool}
\begin{lstlisting}[basicstyle=\ttfamily\scriptsize, breaklines=true, frame=single]
You are a physics reasoning agent in the VIRTUAL TOOL environment.

ENVIRONMENT RULES
- The scene is 256x256 pixels with a white background.
- Objects are colored: black, blue, green, gray.
- Static objects: BLACK is immovable.
- Dynamic objects: GREEN fall under gravity in 2D.
- Block Zone: DARK GRAY ZONE represent that the object can't be placed there
- Scene boundaries are walls/floor.
- Objects shapes: BALL, BAR, JAR, TRIANGLE, TRAPEZOID.
- The goal: make any [GREEN *] object touch any [BLUE *] object.
- You can add exactly one RED OBJECT as the action: [x, y, r]
  - x in [0, 255] (0=left edge, 255=right edge)
  - y in [0, 255] (0=top edge, 255=bottom edge)
  - r in [2, 32]
- The RED OBJECT has shape <RED_OBJECT_SHAPE>.
- you can't place object at the Dark Gray zone.

OBJECT NAMING CONVENTION (STRICT)
- When you mention an object, always use: [COLOR SHAPE] in UPPERCASE,
  e.g. [GREEN BALL], [BLUE BAR], [GRAY JAR], [GRAY TRIANGLE], [BLUE TRAPEZOID].
- The added object must always be written exactly as: [RED OBJECT]
- EXCEPTION: In reasoning sections only, you may use lowercase natural language for readability.

INPUT
This is the input scene: <image>

TASK (what to do)
1) Identify the goal pair: which [GREEN *] must touch which [BLUE *].
2) Identify blockers: walls, gaps, covers, containers, slopes, platforms, and black obstacles.
3) Choose the main causal strategy (PUSH/ROTATE/HIT/TILT/BLOCKED/SPIN/COLLIDE WITH/DEFLECTED/DEFLECT/STOPPED/SUPPORT/BLOCK/STOP/STRIKE).
4) Estimate coordinates of relevant objects (approximate is OK but must be plausible).
5) Describe the red object placement relative to objects AND the whole scene.
6) Output the final action [x, y, r].
7) You are encouraged to propose different reasoning in each section and different strategies to improve diversity.

OUTPUT FORMAT (MUST MATCH EXACTLY)
You MUST output EXACTLY these sections in EXACT order, and NOTHING else:

<scene_reasoning>
...
</scene_reasoning>
<scene_answer>
...
</scene_answer>
<causal_actions_reasoning>
...
</causal_actions_reasoning>
<causal_actions_answer>
...
</causal_actions_answer>
<placement_reasoning>
...
</placement_reasoning>
<placement_answer>
...
</placement_answer>
<action>
[x, y, r]
</action>

CONTENT RULES PER SECTION

<scene_reasoning>
REASONING REQUIREMENTS:
- Identify all relevant objects and estimate their approximate positions.
- State the goal clearly (which green touches which blue).
- Analyze spatial relationships: alignment, distance, and location relative to the scene center.
- Identify obstacles between goal objects.
- Assess whether objects will naturally interact or need intervention.
</scene_reasoning>

<scene_answer>
- List the key goal objects (at least the [GREEN *] and its goal [BLUE *], good to include [BLACK *]).
- Include any critical blockers or intermediate objects referenced in your reasoning.
- Use THIS exact line style:
  [OBJECT] is at [GLOBAL SPATIAL] [x, y] with size [d].
- GLOBAL SPATIAL must be one of:
  ['TOP-LEFT', 'TOP-RIGHT', 'CENTER', 'TOP', 'BOTTOM-RIGHT', 'BOTTOM-LEFT', 'LEFT', 'BOTTOM', 'RIGHT']
- x, y, d must be integers in range 0-255.
</scene_answer>

<causal_actions_reasoning>
REASONING REQUIREMENTS:
- Reference coordinates from <scene_answer> when describing positions.
- Choose and justify ONE causal strategy.
- Describe the event chain: red object -> interaction -> target motion -> goal achievement.
- Explain the contact point and why it creates the desired motion.
- Verify the approach avoids obstacles.
</causal_actions_reasoning>

<causal_actions_answer>
- Output 1-2 lines total using one of the following templates:

Template Block (for * BALL or * JAR):
  ACTION in ['DEFLECT', 'BLOCK', 'STOP']
  The [RED OBJECT] should [ACTION] the [TARGET OBJECT] at the contact point [x, y].

Template Collision (for * BALL, * JAR, * TRIANGLE, or * TRAPEZOID):
  ACTION in ['DEFLECTED', 'COLLIDE WITH', 'BLOCKED', 'STOPPED', 'PUSH', 'STRIKE', 'HIT']
  DIRECTION in ['RIGHT', 'LEFT']
  The [RED OBJECT] should [ACTION] the [TARGET OBJECT] at the contact point [x, y] to push it towards [DIRECTION].

Template Rotation (for * BAR, * JAR, * TRIANGLE, or * TRAPEZOID):
  ACTION in ['TILT', 'ROTATE', 'SPIN']
  DIRECTION in ['CLOCKWISE', 'COUNTER-CLOCKWISE']
  The [RED OBJECT] should [ACTION] the [TARGET OBJECT] at the contact point [x, y] to rotate it [DIRECTION].

Template Support (for * BAR, * JAR, * TRIANGLE, or * TRAPEZOID):
  The [RED OBJECT] should [SUPPORT] the [TARGET OBJECT] at the contact point [x, y].

For chain actions on the GREEN object:
  Template Collision: The chain actions cause the [GREEN BALL] to move [LEFT|RIGHT] at point [x, y].
  Template Block:     The chain actions cause the [GREEN BALL] to be [STOPPED|DEFLECTED|BLOCKED] at point [x, y].
  Template Rotation:  The chain actions cause the [GREEN BAR] to move [CLOCKWISE|COUNTER-CLOCKWISE] at point [x, y].
</causal_actions_answer>

<placement_reasoning>
REASONING REQUIREMENTS:
- Reference target position from <scene_answer> and contact point from <causal_actions_answer>.
- Calculate red object's initial position to achieve the desired contact point.
- Account for object radius and gravity trajectory.
- Justify radius choice (larger for momentum, smaller for precision).
- CRITICAL PHYSICS RULE: To push an object to a [DIRECTION], the [RED OBJECT] must be placed on the OPPOSITE side.
</placement_reasoning>

<placement_answer>
- Output 1-3 lines using these templates:
  [RED OBJECT] is located at the [RELATIVE SPATIAL] of the [TARGET OBJECT A].
  [RED OBJECT] is located at the [RELATIVE SPATIAL] of the [TARGET OBJECT B].
  [RED OBJECT] is located at the [GLOBAL SPATIAL] of the [WHOLE SCENE].

- RELATIVE SPATIAL in ['TOP-LEFT', 'TOP-RIGHT', 'BOTTOM-LEFT', 'BOTTOM-RIGHT']
- GLOBAL SPATIAL in ['TOP-LEFT', 'TOP-RIGHT', 'CENTER', 'TOP', 'BOTTOM-RIGHT', 'BOTTOM-LEFT', 'LEFT', 'BOTTOM', 'RIGHT']
</placement_answer>

<action>
- Output ONLY: [x, y, r]
- x, y in [0, 255], r in [2, 32]
- Do not add any explanation here.
</action>
\end{lstlisting}

\subsection{CraftVQA}
\label{prompt-Craft-VQA}
\begin{lstlisting}[basicstyle=\ttfamily\scriptsize, breaklines=true, frame=single]
You are a physics reasoning agent in a simulation environment.

ENVIRONMENT RULES
- The scene is 256x256 pixels with a white background.
- Static objects: BLACK and PURPLE are immovable.
- Dynamic objects: GRAY, BLUE, GREEN, CYAN fall under gravity in 2D.
- Scene boundaries are walls/floor.
- Objects shapes: BALL, BAR, JAR, TRIANGLE.
- The goal: Use physical reasoning to predict outcomes and answer the Question.

OBJECT NAMING CONVENTION (STRICT)
- When you mention an object, always use: [COLOR SHAPE] in UPPERCASE, e.g. [GRAY TRIANGLE], [BLUE TRIANGLE], [PURPLE BAR], [BLACK BAR].
- EXCEPTION: In reasoning sections only, you may use lowercase natural language for readability.

INPUT
This is the input scene: <image>
Question: <QUESTION_TEXT>

TASK (what to do)
1) Identify all relevant objects and their approximate positions.
2) Analyze the physical chain of events (gravity, collisions, rotations, and support).
3) Identify blockers: walls, gaps, covers, containers, slopes, and platforms.
4) Evaluate counterfactuals or temporal sequences.
5) Formulate a step-by-step causal reasoning trace.
6) Output the final concise answer.

OUTPUT FORMAT (MUST MATCH EXACTLY)
You MUST output EXACTLY these sections in EXACT order, and NOTHING else:

<scene_reasoning>
- Identify all relevant objects and estimate their approximate positions.
- Analyze spatial relationships: alignment, distance, and location relative to the scene center.
- Identify the potential trajectory of dynamic objects under gravity.
- Address the specific context of the question.
</scene_reasoning>

<scene_answer>
- List the key objects referenced in the question or the physical chain.
- Use THIS exact line style:
  [OBJECT] is at [GLOBAL SPATIAL] [x, y] with size [d].
- GLOBAL SPATIAL must be one of:
  ['TOP-LEFT', 'TOP-RIGHT', 'CENTER', 'TOP', 'BOTTOM-RIGHT', 'BOTTOM-LEFT', 'LEFT', 'BOTTOM', 'RIGHT']
- x, y, d must be integers in range 0-255.
</scene_answer>

<causal_actions_reasoning>
- Reference coordinates from <scene_answer> when describing positions.
- Describe the event chain: object movement -> interaction -> secondary motion -> final state.
- If the question asks about object removal, simulate the scene without those blockers.
- Use causal verbs: PUSH, ROTATE, HIT, TILT, BLOCKED, COLLIDE WITH, DEFLECTED, SUPPORT.
- Justify the conclusion based on the simulated interaction.
</causal_actions_reasoning>

<causal_actions_answer>
- Output 1-2 lines using one of the following templates:

Template Interaction:
  The [OBJECT A] will [ACTION] the [OBJECT B] at point [x, y].

Template Final State:
  The [OBJECT] will end at [GLOBAL SPATIAL] location [x, y] or enter the [CONTAINER/BASKET].

Template Motion:
  The chain actions cause the [OBJECT] to move [DIRECTION] at point [x, y].
</causal_actions_answer>

<final_answer>
- Provide ONLY the direct answer to the question (e.g., "true", "false", "2", "blue").
</final_answer>
\end{lstlisting}




\end{document}